\documentclass{ecai} 

\usepackage{latexsym}
\usepackage{amssymb}
\usepackage{amsmath}
\usepackage{amsthm}
\usepackage{booktabs}
\usepackage{enumitem}
\usepackage{graphicx}
\usepackage{color}
\usepackage{amsfonts,dsfont}
\usepackage{subcaption}
\usepackage{multirow,booktabs}
\usepackage[table]{xcolor}

\newcommand{\BibTeX}{B\kern-.05em{\sc i\kern-.025em b}\kern-.08em\TeX}

\begin{document}


\begin{frontmatter}


\paperid{7065} 


\title{Exploring Topological Bias in Heterogeneous Graph Neural Networks}


\author[A]{\fnms{Yihan}~\snm{Zhang}\thanks{Email: zyh23@mails.tsinghua.edu.cn.}}

\address[A]{Shenzhen International Graduate School, Tsinghua University}




\begin{abstract}
Graph Neural Networks (GNNs) are characterized by their capacity of processing graph-structured data. However, due to the sparsity of labels under semi-supervised learning, they have been found to exhibit biased performance on specific nodes. This kind of bias has been validated to correlate with topological structure and is considered as a bottleneck of GNNs' performance.
Existing work focuses on the study of homogeneous GNNs and little attention has been given to topological bias in Heterogeneous Graph Neural Networks (HGNNs). In this work, firstly, in order to distinguish distinct meta relations, we apply meta-weighting to the adjacency matrix of a heterogeneous graph. Based on the modified adjacency matrix, we leverage PageRank along with the node label information to construct a projection. The constructed projection effectively maps nodes to values that strongly correlated with model performance when using datasets both with and without intra-type connections, which demonstrates the universal existence of topological bias in HGNNs. To handle this bias, we propose a debiasing structure based on the difference in the mapped values of nodes and use it along with the original graph structure for contrastive learning. Experiments on three public datasets verify the effectiveness of the proposed method in improving HGNNs' performance and debiasing.
\end{abstract}

\end{frontmatter}


\section{Introduction}

\label{sec:intro}
In the realm of real-world applications, graph-structured data is ubiquitous, encompassing various domains such as social networks, transportation systems, and biological networks. To effectively analyze and leverage this data, Graph Neural Networks (GNNs) \cite{kipf2017semi,hamilton2017inductive,velivckovic2018graph,xu2018powerful} were developed, providing powerful tools for understanding and extracting insights from homogeneous graph structures. However, many real-world datasets are not homogeneous; they contain heterogeneous data where nodes and edges can have different types, attributes, and relationships, as seen in social platforms with varied user roles and interaction types, or recommendation systems with diverse user-item interactions. This type of data is abstracted as heterogeneous graphs (also known as heterogeneous information networks). To handle such diversity, Heterogeneous Graph Neural Networks (HGNNs) \cite{wang2019heterogeneous,hu2020heterogeneous,fu2020magnn} were introduced to accommodate and exploit the rich diversity of information present in such data, enabling more nuanced and effective modeling in complex systems.

Despite their capabilities, existing GNNs often face an issue known as structural unfairness \cite{tang2020investigating,han2024towards}, which arises when the learned model disproportionately favors certain groups or nodes due to imbalances in the graph’s structure, e.g. the long-tailed power-law distribution \cite{barabasi1999emergence,albert2002statistical}.
Structural unfairness is manifested as nodes in certain regions of a graph—often representing marginalized or minority groups—receiving less accurate predictions or being underrepresented in the learned embeddings, leading to biased outcomes in tasks like recommendation or classification. We refer to this phenomenon as \textit{topological bias} in informal terms.
It has been found to be somewhat correlated with node degree \cite{tang2020investigating} as the nodes with higher degree have a higher probability of being connected to the labeled nodes, and remains as a bottleneck for the overall performance of GNNs.
Based on this discovery, DSGCN \cite{tang2020investigating} proposes a degree-specific GCN layer and utilizes RNN to generate parameters for different degrees sequentially, which improves GNNs' overall performance. However, the limitation is that simply considering node degree discards a large proportion of information about connectivity patterns. LPSL \cite{han2024towards} constructs an advanced PageRank-based debiasing structure, which considers the recursive impact of connected nodes, yielding better performance in mitigating such bias.
\begin{figure}[t]
    \centering
    \includegraphics[width=\linewidth]{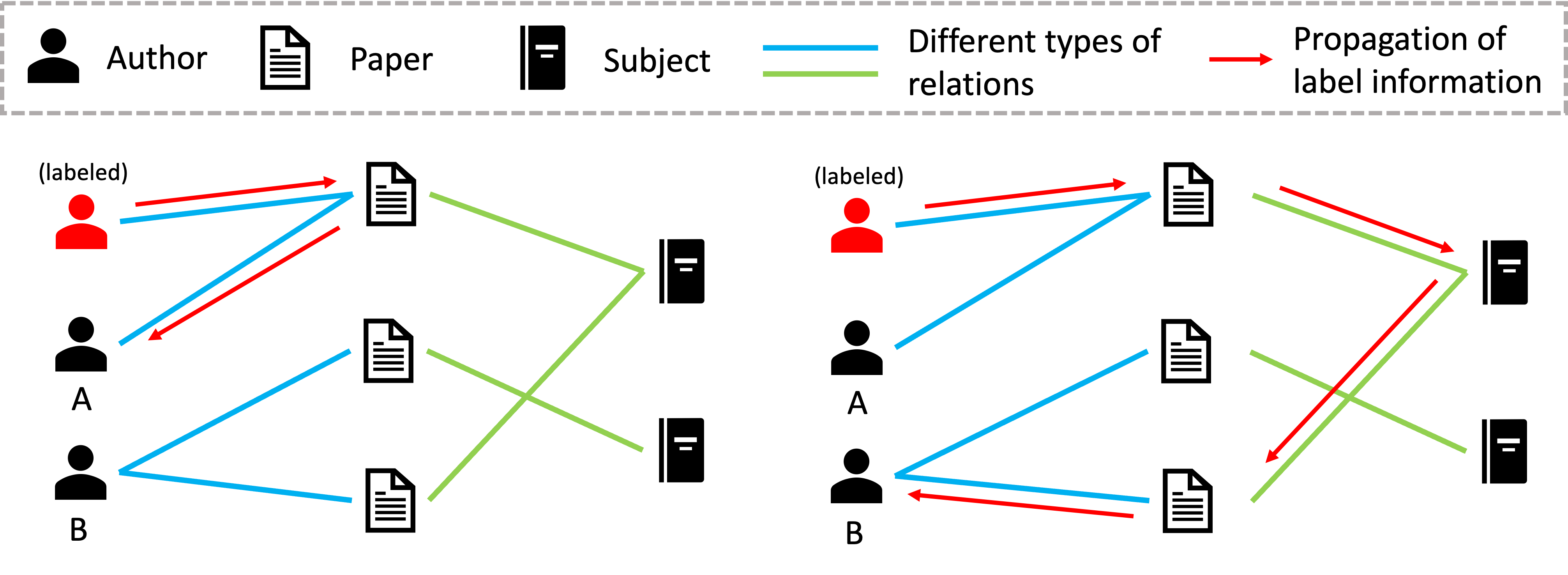}
    \caption{An toy example of topological bias in heterogeneous graphs. Under semi-supervised setting, only limited number of nodes are labeled. Node A is semantically closer to the labeled node illustrated in the figure, thus the message-passing network learns better embeddings for downstream tasks on node A than on node B.}
    \vspace{1.5\baselineskip}
    \label{fig:example}
\end{figure}

While existing work has explored topological bias in homogeneous graphs and provided solutions, there has been no research on heterogeneous graphs yet.
Figure \ref{fig:example} gives a toy example where two authors collaborate to complete an article, one of whose research field is known, while another author B does not collaborate with them on any article.
For general GNN models (including HGNN models), the message passing (or graph convolution) times it takes for the label information to propagate to each node is different. Empirically, we infer that HGNN models tend to provide more accurate inference on node A than node B as node A is semantically closer to the labeled node.
This work aims to investigate and exploit such kind of performance discrepancies of HGNNs rooted in the topological structure. Firstly, we propose that examining solely from the perspective of node degree is insufficient for revealing this phenomenon. Specifically, we construct a meta-weighted graph which seperately consider intra-type and inter-type connections in order to enhance the semantics of the original graph. Then we construct a projection based on the meta-weighted graph which maps nodes to values that strongly correlated with the performance of the HGNN models, and the experiment results present that topological bias also inherently exists in HGNNs.
In order to mitigate such bias, we propose a debiasing structure which leverages the difference in mapped values between nodes and adopt graph contrastive learning to learn representations under semi-supervised condition.



Our main contributions are as follows:
\begin{itemize}
    \item We first investigate topological bias in HGNNs and verify its existence by constructing an effective projection.
    \item 
    We design a novel heterogeneous graph constrastive learning method based on the constructed projection to mitigate the bias observed in HGNNs under semi-supervised scenario.
    \item Experiments on three public datasets demonstrate the effectiveness of the proposed debiasing method -- HTAD, and additional studies further verify the efficiency and robustness of HTAD.
\end{itemize}

\section{Preliminaries and Related Work}
\subsection{Heterogeneous Graphs}
We denote a graph as \(\mathcal{G}=(\mathcal{V},\mathcal{E};\mathcal{T})\), where \(\mathcal{V}\) is the set of nodes, \(\mathcal{E}\) is the set of edges and \(\mathcal{T}\) is the set of node types.
A graph \(\mathcal{G}\) is considered heterogeneous when \(|\mathcal{T}|\ge2\), i.e. there are at least two different types of nodes \cite{dong2017metapath2vec}.
The total number of nodes is denoted as \(N=|\mathcal{V}|\), and the adjacency matrix can be represented as \(\mathbf{A}\in\mathbb{R}^{N\times N}\). In specific, for a heterogeneous graph, we sort the nodes by type and use this sorted node list as the index order for the adjacency matrix.
The set of node feature matrices is denoted as \(\mathbf{X}=\{\mathbf{X}_t \in \mathbb{R}^{N_t \times d_t} | t \in \mathcal{T}\}\), where \(N_t\) and \(d_t\) are the number and feature dimension for nodes of type \(t\), respectively. Let \(\mathbf{Y} \in \{0,1\}^{N_{t^*} \times C}\) be the label matrix, where \(t^*\) is the target type and \(C\) is the number of classes. \(\mathbf{Y}_{ij} = 1\) indicates that node \(i\) belongs to category \(j\), while \(\mathbf{Y}_{ij} = 0\) otherwise. 
Edges in a heterogeneous graph can be categorized as either \textit{intra-type} or \textit{inter-type} connections.
A heterogeneous graph without intra-type connections can be regarded as a multipartite graph, meaning that nodes of each type form an independent set.
\textit{Meta relations} refer to the relationships between different types of nodes which is formulated as \(\mathcal{T}_e\subseteq\mathcal{T}\times\mathcal{T}\), and a \textit{meta-path} is a sequence of node types and edge types that defines a specific, meaningful connection pattern. For example, in a bibliographic graph, a meta-path like Author → Paper → Author represents co-authorship, capturing the relationship that two authors are connected via a paper they wrote together.
In real-world heterogeneous graphs, meta-relations are typically sparse, meaning that the non-zero blocks in the adjacency matrix are limited and scattered.

\subsection{Heterogeneous Graph Neural Networks (HGNNs)}
Standard GNNs (like GCN, GAT) treat all nodes and edges the same, which is not effective for heterogeneous graphs where different types of nodes and relations have different meanings.
HGNNs solve these challenges by using type-specific transformations, relation modeling, and meta-paths to learn more meaningful node embeddings.
Existing HGNNs can be roughly divided into two categories, respectively \textit{meta-path based} and \textit{meta-path free} \cite{yang2023simple}.
Based on pre-defined meta-paths, HAN \cite{wang2019heterogeneous} performs a two-level attention mechanism to learn the weight of meta-path based node pairs and capture semantic relations between distinct meta-paths respectively. 
Instead of pre-defining meta-paths, HGT \cite{hu2020heterogeneous} which is a Transformer-based architecture directly parameterizes each edge based on its meta relation triplet and integrates the parameter into the calculation of attention.
\cite{lv2021we} improves generic GNNs to perform well on heterogeneous graphs and proposes simple-HGN which employs residual connections for updating both node representations and edge attention scores, and normalization for the output embeddings.

\subsection{Graph Contrastive Learning (GCL)}
Contrastive learning \cite{chen2020simple} is rooted in the principle of learning representations by maximizing the agreement between similar instances while minimizing agreement between dissimilar instances. In GCL, this principle is adapted to graph-structured data, where nodes, edges, and subgraphs serve as instances \cite{zhu2021an}.
Inspired by Deep InfoMax \cite{hjelm2018learning,bachman2019learning}, DGI \cite{velivckovic2018deep} introduces contrastive learning to GNNs, which applies a corruption function on the original graph to obtain a negative sample and constructs optimization objective based on maximizing mutual information between different graph views.
The corruption function can be interpreted as \textit{topology-based augmentations} which involves modifications to the graph structure, such as node dropping, edge perturbation, or subgraph sampling \cite{zhu2021an,zhu2020deep,zhu2021graph,hassani2020contrastive}.
After obtaining the augmented graph, \textit{node-level contrast} learns discriminative representations for individual nodes by contrasting their embeddings across different augmented views.
Based on the original multual information maximization method, GraphCL \cite{you2020graph} improves the contrastive loss function by incorporating similarity between graph representations.
Under semi-supervised learning scenario, GCL can benefit from the label information by combining supervised label loss with unsupervised contrastive loss \cite{wan2021contrastive}.
GCL can also be combined with degree-related bias to learn better representations, e.g. GRADE \cite{wang2022uncovering} proposes an edge perturbation scheme based on the degree disparities between nodes to generate augmentation.


There are also heterogeneous graph contrastive methods under self supervised \cite{park2020unsupervised,jiang2021pre,che2021multi,zhu2022structure,chen2023heterogeneous,yu2024heterogeneous} and semi-supervised settings \cite{yao2023semi}. Adapted to heterogeneous graphs, the contrast can be performed between global view and meta-path based view \cite{wang2021self,jiang2021contrastive} to enhance the expressiveness of the model.


\begin{figure*}
    \centering
    \begin{subfigure}{0.15\linewidth}
        \centering
        \includegraphics[width=\linewidth]{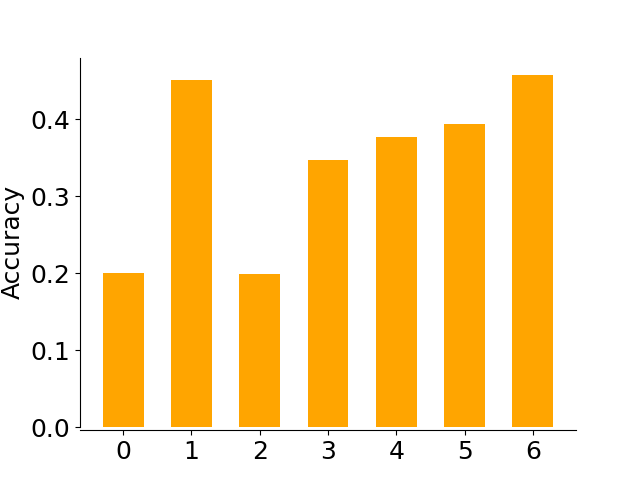}
        \caption{\centering Degree, 0.1, \(r_s=0.4286\)}
    \end{subfigure}
    \hfill
    \begin{subfigure}{0.15\linewidth}
        \centering
        \includegraphics[width=\linewidth]{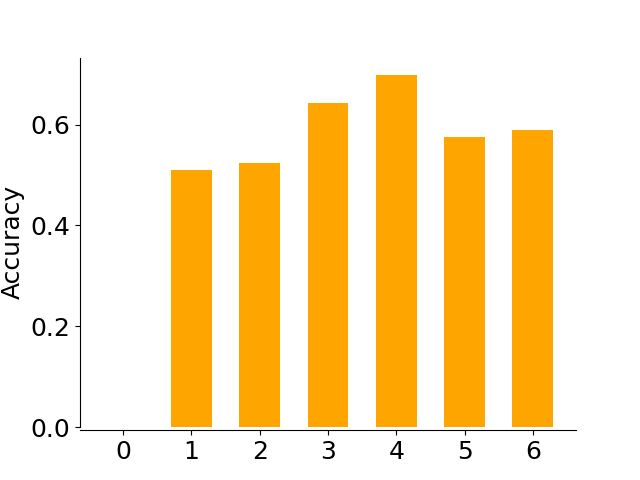}
        \caption{\centering Degree, 0.5, \(r_s=0.7143\)}
    \end{subfigure}
    \hfill
    \begin{subfigure}{0.15\linewidth}
        \centering
        \includegraphics[width=\linewidth]{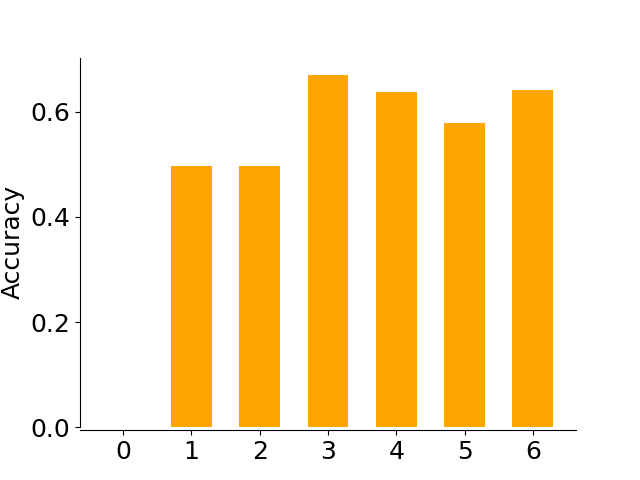}
        \caption{\centering Degree, 1.0, \(r_s=0.75\)}
    \end{subfigure}
    \hfill
    \begin{subfigure}{0.15\linewidth}
        \centering
        \includegraphics[width=\linewidth]{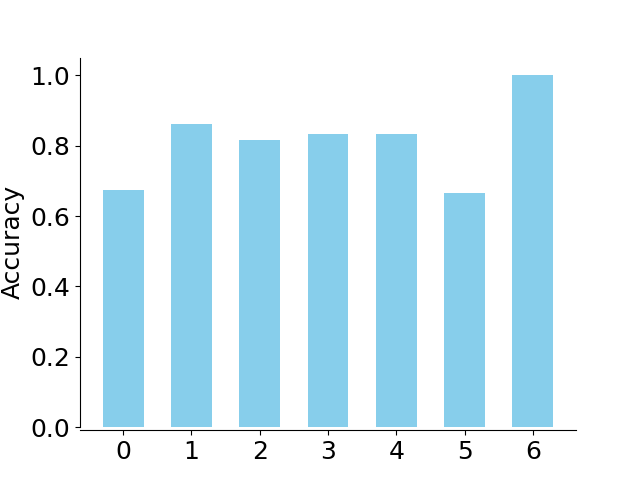}
        \caption{\centering Degree, 0.1, \(r_s=0.25\)}
    \end{subfigure}
    \hfill
    \begin{subfigure}{0.15\linewidth}
        \centering
        \includegraphics[width=\linewidth]{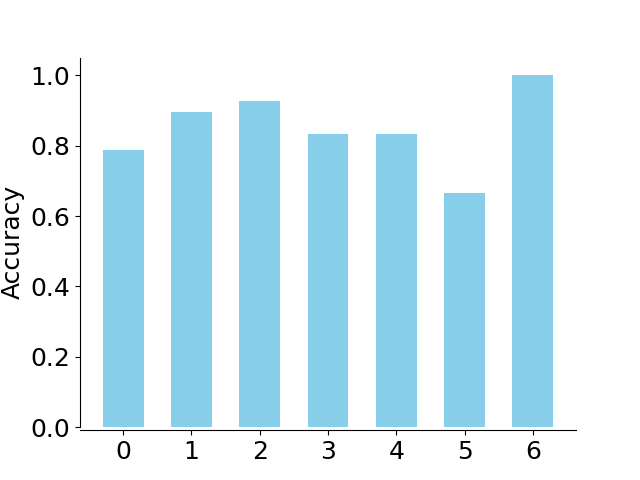}
        \caption{\centering Degree, 0.5, \(r_s=0.1429\)}
    \end{subfigure}
    \hfill
    \begin{subfigure}{0.15\linewidth}
        \centering
        \includegraphics[width=\linewidth]{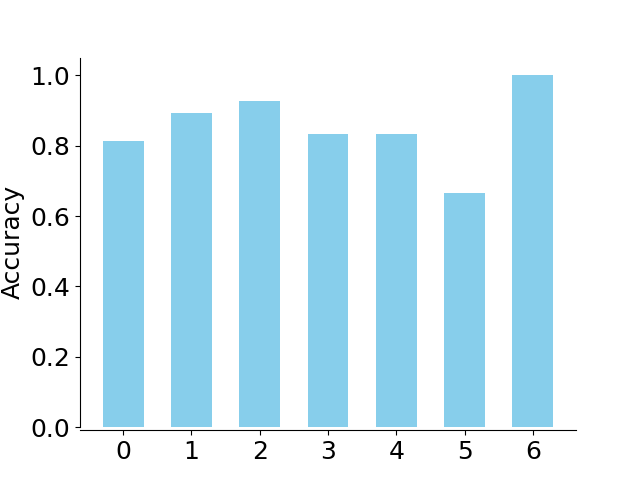}
        \caption{\centering Degree, 1.0, \(r_s=0.1429\)}
    \end{subfigure}
    \vspace{\baselineskip}
    
    \begin{subfigure}{0.15\linewidth}
        \centering
        \includegraphics[width=\linewidth]{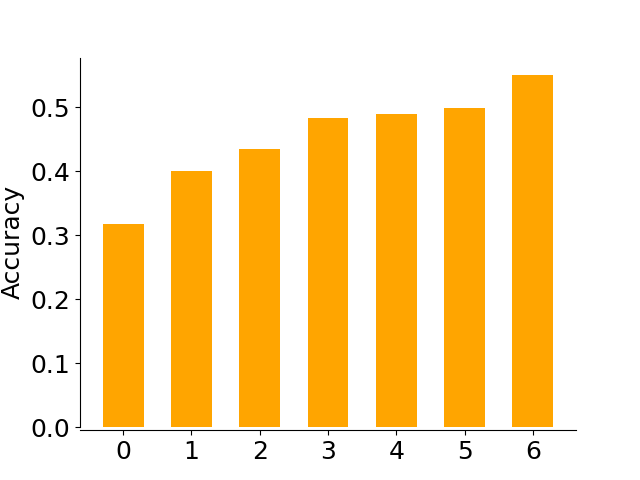}
        \caption{\centering HLID, 0.1, \(r_s=1.0\)}
    \end{subfigure}
    \hfill
    \begin{subfigure}{0.15\linewidth}
        \centering
        \includegraphics[width=\linewidth]{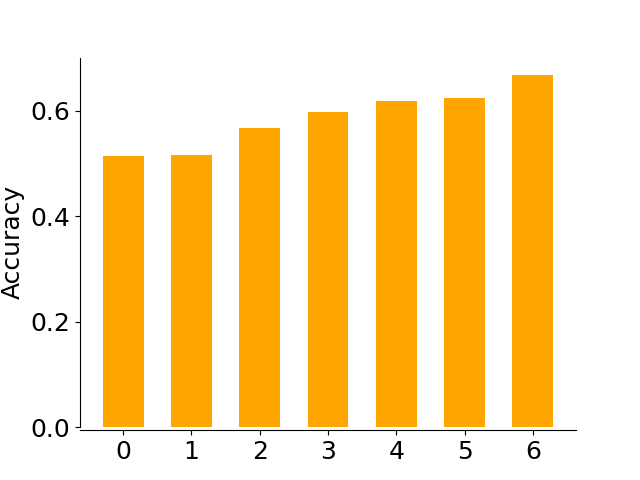}
        \caption{\centering HLID, 0.5, \(r_s=1.0\)}
    \end{subfigure}
    \hfill
    \begin{subfigure}{0.15\linewidth}
        \centering
        \includegraphics[width=\linewidth]{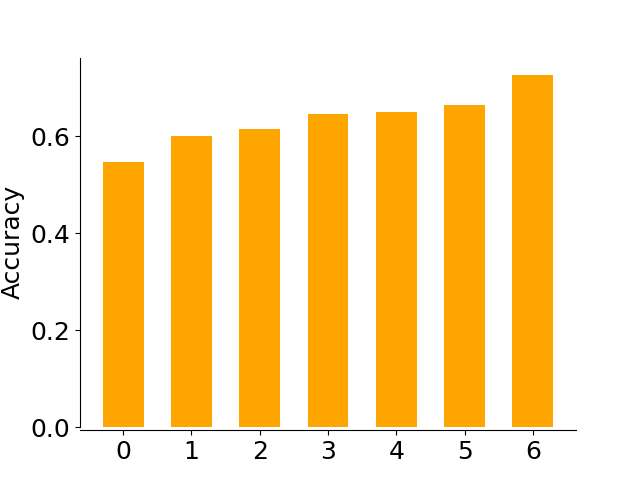}
        \caption{\centering HLID, 1.0, \(r_s=1.0\)}
    \end{subfigure}
    \hfill
    \begin{subfigure}{0.15\linewidth}
        \centering
        \includegraphics[width=\linewidth]{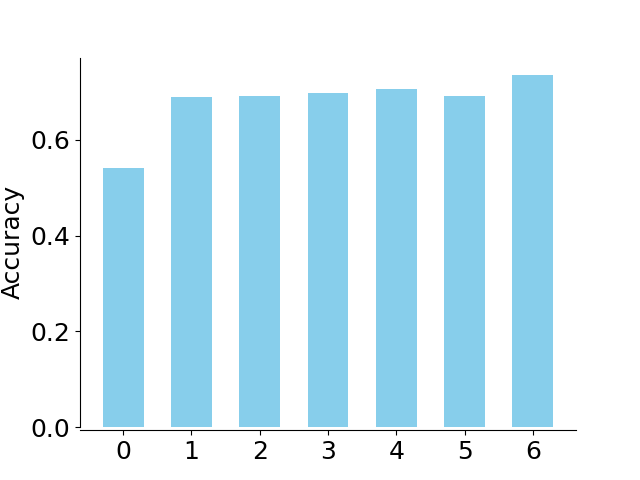}
        \caption{\centering HLID, 0.1, \(r_s=0.8929\)}
    \end{subfigure}
    \hfill
    \begin{subfigure}{0.15\linewidth}
        \centering
        \includegraphics[width=\linewidth]{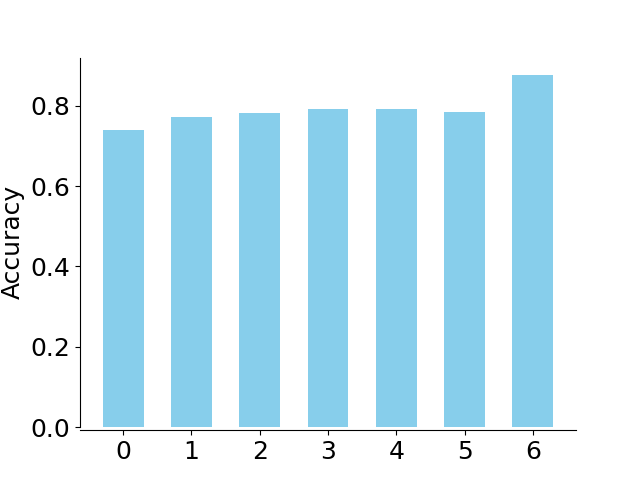}
        \caption{\centering HLID, 0.5, \(r_s=0.8571\)}
    \end{subfigure}
    \hfill
    \begin{subfigure}{0.15\linewidth}
        \centering
        \includegraphics[width=\linewidth]{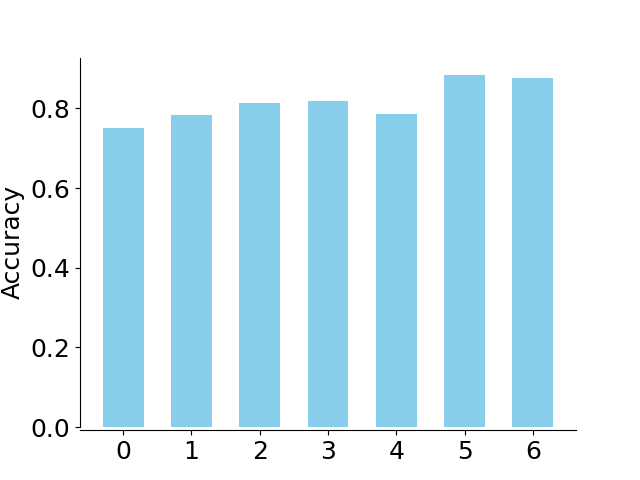}
        \caption{\centering HLID, 1.0, \(r_s=0.8571\)}
    \end{subfigure}
    \vspace{\baselineskip}
    \caption{Prediction accuracy under different projections (projection/label rate). The horizontal axis is bucket id. The orange ones are the results on IMDB dataset, and the blue ones are DBLP. The base model is HAN and number of buckets is set to 7. Best viewed in color.}
    \label{fig:bias}
\end{figure*}

\begin{figure}
    \centering
    \begin{subfigure}{0.32\linewidth}
        \centering
        \includegraphics[width=\linewidth]{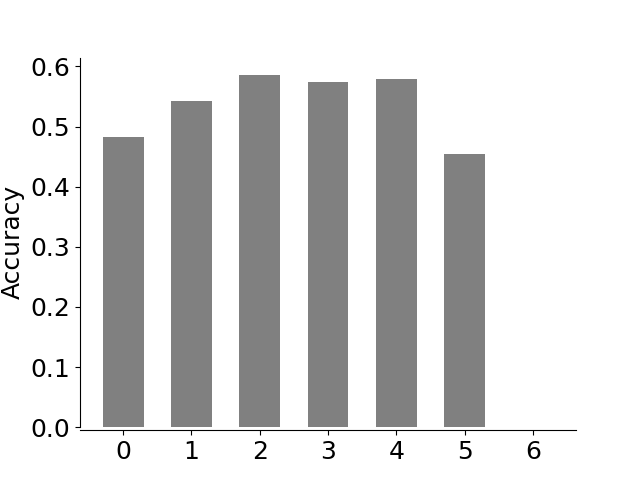}
        \caption{\centering Degree, \(r_s=-0.3929\)}
    \end{subfigure}
    \hfill
    \begin{subfigure}{0.32\linewidth}
        \centering
        \includegraphics[width=\linewidth]{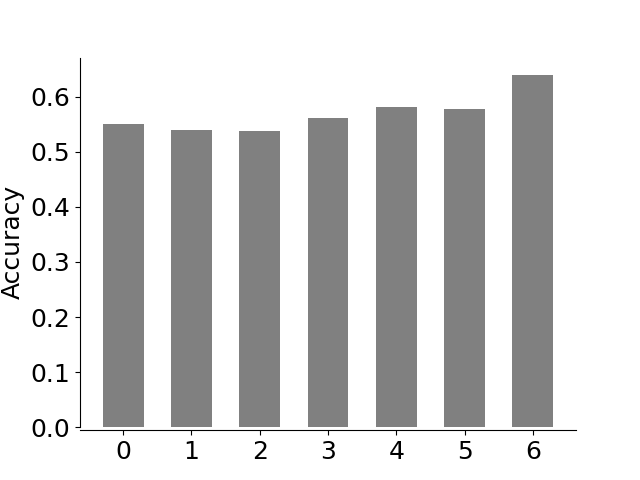}
        \caption{\centering HLID (w/o \(\eta_1\)- term), \(r_s=0.8214\)}
        \label{fig:wo}
    \end{subfigure}
    \hfill
    \begin{subfigure}{0.32\linewidth}
        \centering
        \includegraphics[width=\linewidth]{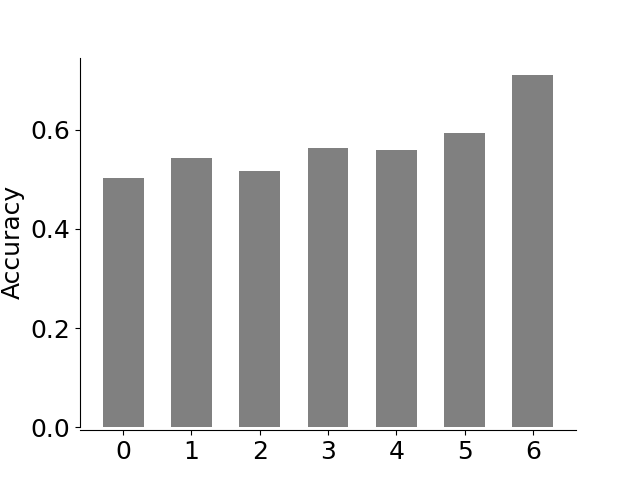}
        \caption{\centering HLID, \(r_s=0.9286\)}
        \label{fig:mw}
    \end{subfigure}
    \vspace{\baselineskip}
    \caption{Prediction accuracy under different projections. The dataset used has intra-type connections (ACM), and the label rate is 0.01.}
    \vspace{1.5\baselineskip}
    \label{fig:acm}
\end{figure}


\section{Topological Bias in Heterogeneous Graph Neural Networks} \label{sec:bias}
In this section, we first discuss how to measure the quality of a projection on graph nodes, and then contruct an effective projection based on the meta-weighted graph. Finally, we experimentally verify the existence of topological bias in heterogeneous graphs using the constructed projection.


\subsection{Projection Quality Measurement}
In order to explore topological bias on heterogeneous graphs, it is essential to find an appropriate projection which maps the nodes in the graph to meaningful metrics. This raises the question of how to evaluate the quality of a projection. A `good' projection should correspond to a feature space where the arrangement of nodes presents a certain regularity.
Formally, we define the set composed of all projections \(\mathcal{P}=\{f|f(\cdot): \mathcal{V}\rightarrow\mathbb{R}\}\) where each element maps nodes in a graph to real numbers. The widely adopted projection is node degree, which can be formulated as \(\mathrm{deg}(u)=\sum_{v\in\mathcal{V},v\neq u}\mathbf{A}_{uv}\), where \(\mathrm{deg}\in \mathcal{P}\).
In order to evaluate the quality of \(f\in\mathcal{P}\), we leverage Spearman's rank correlation coefficient \cite{spearman1987proof} which measures the monotonic relationship between two variables to quantify it. Specifically, we calculate an index as
\begin{equation}
    r_s=1-\frac{6\sum_{i=1}^{N}d_i^2}{N(N^2-1)}\in[-1,1],
\end{equation}
where \(d_i=R(f(i))-R(\text{acc}_i)\) is the difference between the ranks of \(f(i)\) and prediction accuracy of a model belonging to a specific category (e.g. message passing neural networks) on \(i\) among all nodes.
The larger the value, the higher the correlation between \(f\) and accuracy, 
and \(f\) should be considered as a better projection.

\subsection{Meta-weighted Graph}
In real-world datasets, connections in heterogeneous graphs usually contains merely a single semantic information, which is manifested as the binary adjacency matrix. However, using this form of matrix for further representation will result in loss of distinct meta-relations information which is essential for heterogeneous graphs.
Therefore, we propose to attach weights to edges according to different meta-relations.
Specifically, we modify the adjacency matrix as follows
\begin{equation}
    \mathbf{B}\gets\mathbf{A}\odot [\mathbf{R}_{ij}\cdot \mathbf{1}^{N_i\times N_j}]_{|\mathcal{T}|\times|\mathcal{T}|},
\end{equation}
where \(\odot\) is Hadamard product, which is equivalent to adding mask blocks to the adjacency matrix. \(\mathbf{R}\in\mathbb{R}^{|\mathcal{T}|\times|\mathcal{T}|}\) is the symmetric type relation matrix, and \(\mathbf{R}_{ij}\) is positively correlated with the intensity of meta-relations \((i,j)\). In particular, when \(\mathbf{R}\) is all-1, \(\mathbf{A}\) keeps unchanged. On this basis, we construct two terms respectively for \textit{self-amplification} and \textit{relation regulation}, which can be formulated as
\begin{equation} \label{eq:mw}
\mathbf{R}_{ij}=1+\underbrace{\eta_1\mathds{1}[i=j]}_{\text{self-amplification}}+\underbrace{\eta_2/\sum_{u\in\mathcal{V}_i}\sum_{v\in\mathcal{V}_j} \mathds{1}[(u,v)\in\mathcal{E}]}_{\text{relation regulation}},
\end{equation}
where \(\eta_1\), \(\eta_2\) are positive scale parameters, \(\mathcal{V}_i\) represents the set of nodes of type \(i\), and \(\mathds{1}[\cdot]\) is the indicator function whose value is 1 if condition holds, 0 otherwise.
The \(\eta_1\)-regulated term is intended to amplify self relation intensity by attaching greater value to intra-type connections, since nodes of the same type can communicate within one hop via intra-type connections while at least two hops through inter-type connections.
Meanwhile, inspired by widely adopted degree normalization technique \cite{kipf2017semi}, we construct the \(\eta_2\)-regulated term in Equation \ref{eq:mw} in order to differentiate meta relations by number of edges. 
After the same number of rounds of message passing, the contributions of meta relations with fewer edges are more concentrated, thus they are assigned higher weights.
The effectiveness of the construction will be validated  in the later discussion.

\subsection{Heterogeneous Label Impact Degree}
Inspired by personalized PageRank \cite{gasteiger2018predict}, in which each element of the propagation matrix represents the influence intensity of one node on another node during the process of message passing, we construct an impact matrix as
\begin{equation}
    \mathbf{Q}=\alpha(\mathbf{I}-(1-\alpha)\Tilde{\mathbf{D}}^{-1/2}\Tilde{\mathbf{B}}\Tilde{\mathbf{D}}^{-1/2})^{-1},
\end{equation}
where \(\alpha\in(0,1]\) is the teleport probability, \(\Tilde{\mathbf{B}}=\mathbf{I}+\mathbf{B}\) is the adjacency matrix  with self-loops, and \(\Tilde{\mathbf{D}}_{ii}=\sum_j\Tilde{\mathbf{B}}_{ij}\). Note that \(\mathbf{B}\) is the meta-weighted adjacency matrix and if \(\mathbf{A}\) is symmetric, \(\mathbf{Q}\) is still symmetric, i.e. the mutual influence between two nodes is identical. 
In order to quantify the impact of all labeled nodes on a given node, we calculate the \textbf{H}eterogeneous \textbf{L}abel \textbf{I}mpact \textbf{D}egree (HLID) as \(\mathbf{Z}=\mathbf{Q}\mathbf{J}\) where \(\mathbf{J}\in\{0,1\}^{N\times 1}\) is the label matrix and \(\mathbf{J}_{i,0}=1\) indicates node \(i\) is labeled, otherwise unlabeled, and the projection can be represented as \(\mathrm{HLID}(i)=\mathbf{Z}_{i,0}\).
The construction guarantees that all sources of influence are of equal strength and decay only with increasing hop count, independent of the degree of the source node.
HLID quantifies the impact of all labeled nodes on each node, which intuitively correlated with model performance on these nodes.


\subsection{Experiment Validation} 
Firstly, we conduct experiments on IMDB and DBLP dataset, which do not have intra-type connections.
Since there is a significant variation in both node degree and HLID across different nodes, we construct \textit{buckets} of specific quantity, sorting nodes according to their projected values and grouping those with similar values into the same bucket, to enhance the clarity of our results.
The experiments are conducted at different label rates (Figure \ref{fig:bias}),
and the accuracy of each bucket is calculated as the average prediction accuracy of the nodes in the bucket.
It's shown that node degree does not effectively capture topological bias in HGNNs, while HLID generally has large \(r_s\) values, demonstrating superior performance across both datasets and varying label rates.
In addition, when the label rate is higher—meaning the labeled nodes are denser—the bias among different nodes becomes less pronounced, as the expected shortest path length from a node to a labeled node decreases.

The self-amplification is only activated when there exist intra-type connections.
As shown in Figure \ref{fig:acm}, HLID without \(\eta_1\)-regulated term performs poorly on ACM dataset
(Figure \ref{fig:wo}), while HLID outperforms it (Figure \ref{fig:mw}), as the projected values showing stronger positive correlation with prediction accuracy (i.e. \(r_s\) is larger).



\begin{figure}[t]
    \centering
    \includegraphics[width=\linewidth]{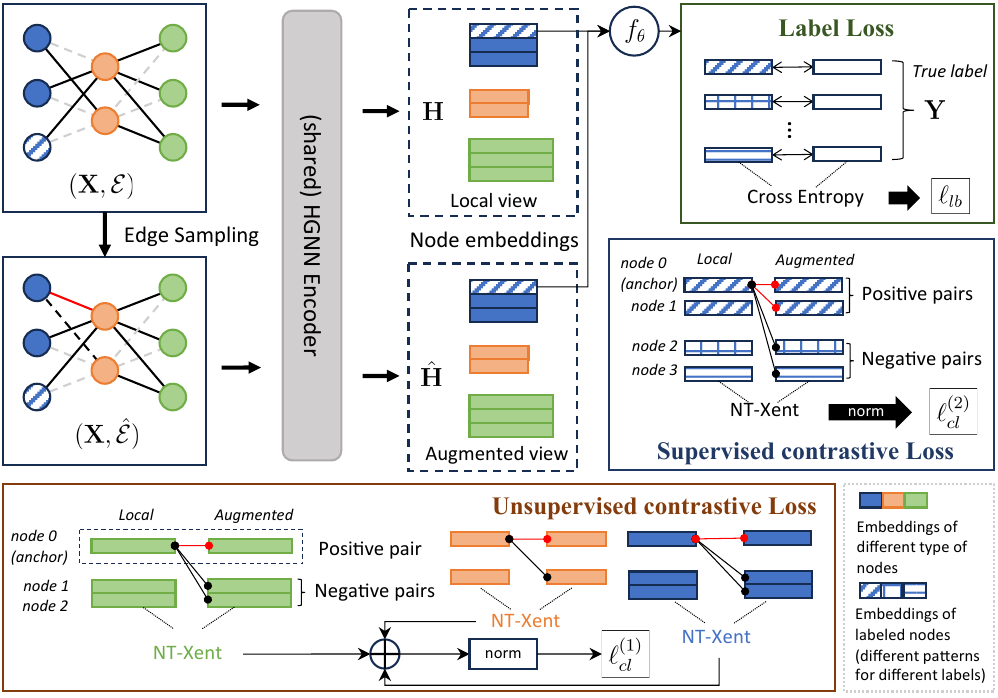}
    \caption{Debiasing method framework diagram.}
    \vspace{\baselineskip}
    \label{fig:fw}
\end{figure}

\section{Debiasing Methodology}
The previous section has already shown the existence of topological bias in HGNNs.
As the discrepancies in HLID metrics has much to do with topological bias, we propose a debiasing framework based on the HLID differences between nodes to mitigate this issue.
Next, we introduce a collaborative training loss that combines two contrastive losses tailored for heterogeneous graphs with the label loss.
The proposed method can be seamlessly integrated into existing HGNNs (and other generic GNNs).
The method framework is depicted in Figure \ref{fig:fw}.



\subsection{Heterogeneous Topological-Aware Debiasing} \label{sec:htad}
Graph contrastive learning has been shown to effectively mitigate degree-related bias in homogeneous graphs by generating additional edges for low-degree nodes (i.e. minority groups), thereby enabling better representation learning on these nodes \cite{wang2022uncovering}.
Inspired by this idea, we perform data augmentation by constructing a \textbf{H}eterogeneous \textbf{T}opological-\textbf{A}ware \textbf{D}ebiasing (HTAD) structure
\(\hat{\mathcal{G}}=(\mathcal{V},\hat{\mathcal{E}})\) based on the original graph structure, where \(\hat{\mathcal{E}}\subseteq\mathcal{V}\times\mathcal{V}\).
When considering edge sampling strategy, we incorporate the HLID projection discussed in the previous section to achieve topological-aware debiasing in heterogeneous graphs.

Firstly, we constrain the sampling space.
If there are additional edges among \(\bar{\mathcal{E}}\) which is the edge set of \(\mathcal{G}\)'s complement graph, i.e. \(\hat{\mathcal{E}}\subseteq\mathcal{E}\cup\bar{\mathcal{E}}\), cardinality of \(\hat{\mathcal{E}}\) could reach \(\mathcal{O}(N^2)\), which implies that both computational and storage overhead increase dramatically. Therefore, we restrict that \(\hat{\mathcal{E}}\) should have the same edge type as \(\mathcal{E}\), i.e. the meta relations set is consistent, and denote the sample space as \(\mathcal{E}_s\) which satisfies \(|\mathcal{E}_s|=\Tilde{\mathcal{O}}(|\mathcal{E}|)\ll \mathcal{O}(N^2)\).
Secondly, we adopt exponential distribution to sample edge weights which serves as a benchmark for weighted random graphs \cite{van2014random}, i.e. \(w_{uv}\sim E(\lambda)\). On this basis, in order to obtain a binary graph, edges with weights lower than a topological attribute value \(\varphi_{uv}\) are retained and the rest are discarded. Then the probability of being reserved is \(\Pr(w_{uv}\le\varphi_{uv})=F(\varphi_{uv})\) where \(F(t)=1-e^{-\lambda t}\) is the cumulative density function (CDF).
Thirdly, instead of i.i.d. sampling, we adopt different sampling strategies for connected and unconnected node pairs in \(\mathcal{G}\), and take into account the HLID value at both endpoints of a specific edge.
Specifically, the probability of \((u,v)\in\hat{\mathcal{E}}\) is formulated as
\begin{equation} \label{eq:pr}
    \hat{p}_{uv}=\begin{cases}
    1-(1-p_0)\exp(-\lambda|\delta(u,v)|),& (u,v)\in \mathcal{E} \\
    1-\exp(-\lambda|\delta(u,v)|),& (u,v)\in \mathcal{E}_s\setminus\mathcal{E}
    \end{cases},
\end{equation}
where \(\delta(u,v)=\mathrm{HLID}(u)-\mathrm{HLID}(v)\) serves as a threshold and \(\lambda\) is the scale parameter which adjusts the density of \(\hat{\mathcal{G}}\). As \(\lambda\) grows, \(\hat{p}_{uv}\) grows, and \(\hat{\mathcal{G}}\) becomes denser.
In order to preserve partial semantics of the original graph, we set a hyperparameter \(p_0\) which ensures that the probability of original edges being reserved is lower bounded by a constant, i.e. \(\hat{p}_{uv}\ge p_0\).
The construction implies that node pairs with greater differences in projected values are more likely to establish edges in \(\hat{\mathcal{G}}\), which increases the likelihood that low-HLID nodes connect to labeled nodes with high HLID values in \(\hat{\mathcal{G}}\).
The newly constructed perturbation edges facilitate message passing between the nodes, providing additional information to support the disadvantaged nodes in learning better representations.

\subsection{Contrastive Loss Function}
Previous GCL methods calculate pairwise loss between each entry of the encoded node embeddings of the original graph and the contrastive graph \cite{zhu2021graph,wan2021contrastive}, or encode the entire graph and calculate loss between pairs of graph embeddings after multiple samplings \cite{you2020graph}.
However, the time complexity of computing the loss between each pair of nodes is \(\mathcal{O}(N^2)\), which is impractical for large graphs.
Additionally, the unified encoding approach may obscure node type information, which is not suitable for heterogeneous graphs.
To address this, we propose a contrastive loss specific for heterogeneous graphs which avoid calculation of a large amount of redundant node pairs, thereby improving efficiency.

In each epoch of the pre-training process, we construct an augmented structure \(\hat{\mathcal{G}}\) using the strategy outlined in the previous section. Node embeddings under local and augmented views, denoted as, \(\mathbf{H}\) and \(\hat{\mathbf{H}}\), are then obtained by inputting \(\mathcal{G}\) and \(\hat{\mathcal{G}}\) (along with node features) into an HGNN-based encoder. 
Based on these two independent views, we construct contrastive loss that consists of two components: a general loss and a target-specific loss.
The general loss is computed separately for each node type.
For each node \(u\), \(\mathbf{h}_u\) and \(\hat{\mathbf{h}}_u\) form a positive pair, while \(\mathbf{h}_u\) and \(\hat{\mathbf{h}}_v\) (where \(v\) is of the same type as \(v\) and \(u\neq v\)) form negative pairs.
We adopt NT-Xent \cite{sohn2016improved} to construct loss based on previously annotated pairs and the total loss normalized by the number of all nodes is formulated as
\begin{equation} \label{eq:cl1}
    \ell_{cl}^{(1)}=-\frac{1}{N}\sum_{t\in\mathcal{T}}\sum_{u\in\mathcal{V}_t} \log\frac{\exp(S(\mathbf{h}_u,\hat{\mathbf{h}}_u))}{\sum_{v\in\mathcal{V}_t,v\neq u} \exp(S(\mathbf{h}_u,\hat{\mathbf{h}}_v))},
\end{equation}
where \(S(\cdot,\cdot)\) is a symmetric, temperature-scaled similarity function, e.g. cosine similarity. This term is an unsupervised loss as no label information is used. By minimizing \(\ell_{cl}^{(1)}\), positive pairs are encouraged to be closer while negative pairs are distinguished.

The target-specific loss is computed only for the nodes of the target type \(t^*\) that have labels. For each anchor node \(u\in\mathcal{V}_{t^*}\), we select \((\mathbf{h}_u,\hat{\mathbf{h}}_v), v\in\mathcal{V}_{t^*}\) as candidate contrastive pairs, and treat the pairs where \(v\) has the same label as \(u\) as positive pairs, otherwise as negative. The loss is then formulated as
\begin{equation} \label{eq:cl2}
    \ell_{cl}^{(2)}=-\frac{1}{N_0}\sum_{u\in\mathcal{V}_{t^*}} \log\frac{\sum_{v\in\mathcal{V}_{t^*},\mathbf{y}_v=\mathbf{y}_u} \exp(S(\mathbf{h}_u,\hat{\mathbf{h}}_v))} {\sum_{v\in\mathcal{V}_{t^*},\mathbf{y}_v\neq\mathbf{y}_u} \exp(S(\mathbf{h}_u,\hat{\mathbf{h}}_v))}.
\end{equation}
The construction takes advantage of the prior knowledge provided by the labels under semi-supervised scenario, which alleviates the adverse impact of false negative pairs to model learning.

\subsection{Overall Loss and Prediction}

We adopt cross entropy loss to measure the difference between the model's output and the true labels. The label loss is then formulated as:
\begin{equation} \label{eq:lb}
    \ell_{lb}=-\sum_{i=1}^{N_0}\sum_{j=1}^{C}\mathbf{Y}_{ij}\log [f_\theta(\mathbf{h}_i, \hat{\mathbf{h}}_i)]_j,
\end{equation}
where \(f_\theta\) is the classifier parameterized by \(\theta\) which maps the node embeddings under two views to a label vector.
Based on the above three losses, the overall loss is constructed as
\begin{equation} \label{eq:overall}
    \mathcal{L}=\ell_{lb}+\lambda_{1}\ell_{cl}^{(1)}+\lambda_{2}\ell_{cl}^{(2)},
\end{equation}
where \(\lambda_{1}>0\), \(\lambda_{2}>0\) are hyperparameters that control the relative importance of two contrastive losses.
Then the model weights can be updated through back propagation.
During the inference stage, we input the original graph structure with node features into the trained HGNN encoder to obtain the node embeddings for downstream tasks.

The main computational overhead comes from the edge sampling procedure which takes \(\mathcal{O}(|\mathcal{E}|)\) and the calculation of two contrastive losses which takes \(\mathcal{O}(\sum_{t\in\mathcal{T}}|\mathcal{V}_t|^2)\). Thus the overall time complexity is \(\mathcal{O}(|\mathcal{E}|+\sum_t|\mathcal{V}_t|^2)\).
Since the construction of the impact matrix only depends on the original graph structure, it can be included in the pre-processing process and will not incur heavy cost.

\begin{table*}[]
\caption{Performance comparison between base models and HTAD enhanced models (micro/macro F1 score, \%)}
    \vspace{\baselineskip}
\label{tab:1}
\centering
\begin{tabular}{cc
>{\columncolor[HTML]{D0CECE}}c c
>{\columncolor[HTML]{D0CECE}}c c
>{\columncolor[HTML]{D0CECE}}c c
>{\columncolor[HTML]{D0CECE}}c c}
\toprule
Dataset                & Label rate & GCN         & GCN+HTAD    & GAT         & GAT+HTAD    & HAN         & HAN+HTAD    & HGT         & HGT+HTAD    \\ \midrule
                       & 5\%        & 77.74/78.13 & 78.64/78.88 & 83.85/84.17 & 85.06/85.44 & 80.57/81.10 & 83.14/83.42 & 64.23/63.49 & 66.34/65.75 \\
                       & 10\%       & 82.22/82.09 & 82.88/82.92 & 86.80/87.28 & 87.86/88.31 & 86.40/86.41 & 87.85/87.87 & 74.18/72.24 & 75.43/74.51 \\
                       & 20\%       & 86.47/86.62 & 86.67/86.79 & 87.86/88.04 & 88.51/88.72 & 89.10/89.27 & 89.82/89.97 & 77.82/77.95 & 78.48/78.65 \\
\multirow{-4}{*}{ACM}  & 50\%       & 88.39/88.63 & 88.48/88.69 & 88.86/89.08 & 89.25/89.48 & 90.24/90.45 & 90.63/90.73 & 82.82/82.98 & 83.03/83.21 \\ \midrule
                       & 5\%        & 50.89/48.91 & 51.20/49.31 & 43.31/41.71 & 44.03/42.11 & 44.86/43.56 & 47.10/45.96 & 48.94/47.25 & 50.15/48.39 \\
                       & 10\%       & 51.92/50.83 & 52.07/50.99 & 47.68/45.76 & 48.11/46.24 & 50.25/48.72 & 52.01/50.36 & 51.68/49.60 & 52.26/50.23 \\
                       & 20\%       & 56.00/54.80 & 56.05/54.93 & 51.35/49.33 & 51.69/49.55 & 55.18/53.32 & 55.77/53.94 & 56.11/53.57 & 56.53/54.05 \\
\multirow{-4}{*}{IMDB} & 50\%       & 59.21/57.76 & 59.23/57.81 & 57.03/54.03 & 57.29/53.93 & 60.98/58.79 & 61.27/59.10 & 61.46/58.79 & 61.67/59.07 \\ \midrule
                       & 5\%        & 61.19/60.58 & 61.92/61.25 & 51.86/51.29 & 52.29/51.79 & 61.74/61.33 & 62.59/62.23 & 57.75/57.46 & 58.79/58.41 \\
                       & 10\%       & 66.18/65.64 & 66.65/66.07 & 57.15/56.01 & 57.70/56.43 & 68.02/67.34 & 68.58/67.85 & 64.71/64.22 & 65.54/65.02 \\
                       & 20\%       & 70.18/69.51 & 70.48/69.80 & 61.07/59.81 & 61.41/60.16 & 72.03/71.36 & 72.51/71.80 & 69.09/68.50 & 69.82/69.32 \\
\multirow{-4}{*}{DBLP} & 50\%       & 73.74/72.96 & 73.92/73.19 & 65.24/64.51 & 65.54/64.72 & 75.93/75.25 & 76.30/75.62 & 72.56/71.83 & 73.13/72.47
\\ \bottomrule
\end{tabular}
\end{table*}

\begin{table*}[]
\centering
\caption{Overall and debiasing performance of different GCL methods with HAN as the base model. For all datasets, label rate is set to 0.05, bucket size 7. (\(\uparrow\) means the larger the better; \(\downarrow\) means the smaller the better; the SOTA results are bolded)
}
\label{tab:debias}
\vspace{\baselineskip}
\begin{tabular}{c|c|ccc|ccc|ccc}
\toprule
\multirow{2}{*}{\begin{tabular}[c]{@{}c@{}}GCL \\ Type\end{tabular}}         & \multirow{2}{*}{Method} & \multicolumn{3}{c|}{ACM}                       & \multicolumn{3}{c|}{IMDB}                      & \multicolumn{3}{c}{DBLP}                       \\ \cmidrule{3-11}
                              & & F1-score $\uparrow$ & tot var $\downarrow$ & bucket var $\downarrow$ & F1-score $\uparrow$ & tot var $\downarrow$ & bucket var $\downarrow$ & F1-score $\uparrow$ & tot var $\downarrow$ & bucket var $\downarrow$ \\ \midrule
None                                & HAN                     & 80.57/81.10          & 0.138          & 0.0433          & 44.86/43.56          & 0.139          & 0.0880          & 61.74/61.33          & 0.199         & 0.0869          \\ \midrule
\multirow{2}{*}{general}            & GRACE                   & 81.97/82.36          & 0.139          & 0.0292          & 45.35/43.83          & 0.142          & 0.0910          & 61.13/60.79          & 0.197         & 0.0831          \\
                                    & GCA                     & 81.73/81.93          & 0.147          & 0.0574          & 45.15/43.43          & 0.145          & 0.0920          & 61.16/60.71          & 0.194         & 0.0814          \\ \midrule
\multirow{5}{*}{\begin{tabular}[c]{@{}c@{}}tailored\\ for \\ HGNNs\end{tabular}} & HeCo                    & 82.04/82.20          & 0.149          & 0.0541          & 46.19/44.39          & 0.138          & 0.0742          & 61.87/61.01          & 0.199         & 0.0762          \\
                                    & STENCIL                 & 80.08/80.54          & 0.157          & 0.0286          & 44.91/43.38          & 0.144          & 0.0918          & 62.13/61.53          & 0.184         & 0.0750          \\
                                    & HGCL                    & 81.56/82.08          & 0.136          & 0.0584          & 46.27/44.61          & 0.139          & 0.0756          & 61.44/60.71          & 0.197         & 0.0923          \\
                                    & MEOW                    & 81.77/81.90          & 0.149          & 0.0562          & 45.48/44.05          & 0.139          & 0.0740          & 61.75/61.14          & \textbf{0.180} & 0.0722          \\
                                    & HTAD                    & \textbf{83.14/83.42} & \textbf{0.131} & \textbf{0.0228} & \textbf{47.10/45.96} & \textbf{0.133} & \textbf{0.0695} & \textbf{62.59/62.23} & 0.182         & \textbf{0.0715} \\ \bottomrule
\end{tabular}
\end{table*}

\begin{table}[]
\centering
\vspace{\baselineskip}
\caption{Quantitative results on node clustering (\%). For all datasets, label rate is set to 0.05. (ACM, DBLP: K-Means; IMDB: FCM)
}
\vspace{\baselineskip}
\label{tab:clus}
\begin{tabular}{c|cc|cc|cc}
\toprule
\multirow{2}{*}{Method} & \multicolumn{2}{c|}{ACM}         & \multicolumn{2}{c|}{DBLP}        & \multicolumn{2}{c}{IMDB}        \\ \cmidrule{2-7}
                        & NMI $\uparrow$ & ARI $\uparrow$ & NMI $\uparrow$ & ARI $\uparrow$ & $l_1 \downarrow$ & $l_2 \downarrow$ \\ \midrule
HAN                     & 50.87          & 52.38          & 15.61          & 9.08           & 40.24          & 48.58          \\
GRACE                   & 53.79          & 57.27          & 16.30          & 8.82           & 40.00          & 48.31          \\
GCA                     & 52.67          & 55.53          & 17.55          & 8.91           & 39.94          & 48.25          \\
HeCo                    & 51.77          & 52.34          & 15.67          & 12.70          & 40.01          & 48.34          \\
STENCIL                 & 51.59          & 49.03          & 15.12          & 9.93           & 40.02          & 48.36          \\
HGCL                    & 50.67          & 46.07          & 15.34          & 9.04           & 39.92          & 48.23          \\
MEOW                    & 55.82          & 61.36          & 15.98          & 10.68          & 40.05          & 48.41          \\
HTAD                    & \textbf{57.94} & \textbf{62.03} & \textbf{18.57} & \textbf{14.36} & \textbf{39.87} & \textbf{48.18}
\\ \bottomrule
\end{tabular}
\end{table}


\section{Experiments}

\subsection{Experiment Settings}
\textbf{Datasets.} We conduct experiments on three heterogeneous graph datasets colleted by \cite{lv2021we}, described as follows:
\begin{itemize}
    \item \textbf{IMDB.} There are four types of entities - Movie (4,932 nodes), Director (2,393 nodes), Actor (6,124 nodes), and Keyword (7,971 nodes). The target type is movie.
    Edge Types: Actor-Movie, Director-Movie, Keyword-Movie.
    Meta-paths: MDM, MAM. This is a multi-label dataset.
    
    \item \textbf{DBLP.} There are four types of entities - Author (4,057 nodes), Paper (14,328 nodes), Term (7,723 nodes), and Venue (20 nodes). The target type is author.
    Edge Types: Author-Paper, Paper-Term, Paper-Venue.
    Meta-paths: APA, APTPA, APVPA.
    
    \item \textbf{ACM.} There are four types of entities - Paper (3,025 nodes), Author (5,959 nodes), Subject (56 nodes), and Term (1,902 nodes).The target type is paper.
    Edge Types: Author-Paper, Paper-Paper, Paper-Subject, Paper-Term.
    Meta-paths: PAP, PSP.

\end{itemize}
Based on the original training set, we further randomly sample a proportion of nodes and retain their labels while removing the labels of other nodes to form the actual training set, in order to validate the robustness of the model to different labeling rates.


\textbf{Baselines.}
Firstly, we select two methods that have shown to be effective for debiasing in homogeneous graphs: \textbf{GRACE} \cite{zhu2020deep} drops edges based on a simple i.i.d. Bernoulli distribution; \textbf{GCA} \cite{zhu2021graph} perform augmentation based on node centrality measures.
We also select GCL methods tailored for HGNNs including: 
\textbf{HeCo} \cite{wang2021self} is an integrated model which has a similar structure to HAN. It constructs two \textit{views} based on node-level attention and semantic-level (meta-path) attention and contrast between them;
\textbf{STENCIL} \cite{zhu2022structure} also utilize these two views and further manually contructs hard negative samples for contrast;
\textbf{HGCL} \cite{chen2023heterogeneous} is recommender system framework, which regards intra-type (user-user, item-item) and inter-type (user-item) connections as two views;
\textbf{MEOW} \cite{yu2024heterogeneous} weights the contrastive terms of negative pairs based on node clustering.
The above methods are originally self-supervised, and we adapt them to semi-supervised learning by incorporating label information.

\textbf{Implementation details and hyperparameters.}
We use Pytorch to conduct all experiments.  Experiments are conducted on Ubuntu 22.04.4 LTS with a single NVIDIA RTX 6000 Ada (48GB) and 256GB RAM. For all HGNN models, we choose the number of message passing layers in \{1, 2\} to avoid over-smoothing, the number of hidden dimensions in \{32, 64, 128\} and the number of attention heads in \{1, 2, 4, 8\} if there is attention mechanism.
Additionally, we apply a GraphNorm layer \cite{cai2021graphnorm} to the output embedding to improve numerical stability.
We use Adam optimizer \cite{kingma2014adam} with learning rate and weight decay searched in \{4e-3, 5e-3, 6e-3\} and \{0.0, 0.001\} respectively. We set the reserve lower bound \(p_0=0.5\) and temperature scale to 1. In order to mitigate the impact of randomness, we select seed from \{0, 1, 42\} and search the optimal parameter space. Then we repeat the experiment 10 times to report the average results.

\subsection{Main Results and Analysis}
We evaluate the effectiveness of the proposed method on two downstream tasks, respectively node classification and node clustering.
In addition, we provide qualitative analysis through visualization.

\textbf{Node classification.}
We apply the proposed HTAD to four GNN models, respectively GCN \cite{kipf2017semi}, GAT \cite{velivckovic2018graph}, HAN \cite{wang2019heterogeneous} and HGT \cite{hu2020heterogeneous}. For GCN and GAT, we adapt it for heterogeneous graphs by applying homogeneous graph convolution on each edge type.
We train each model for 50 epochs which ensures the convergence of all models.
To generate binary predictions, we apply a unit step function to the final embeddings, setting each element to 1 if it exceeds a threshold of 0.5, and to 0 otherwise.
Then we use the widely-adopted F1 score to measure the quality of the results.
Table \ref{tab:1} shows the performance comparison between base models and HTAD enhanced models, from which we have the following findings: (1) Despite the models' sensitivity to label rates, HTAD can generally improve their performance across different datasets, especially when the base model struggles with low label rates, such as HGT on the ACM dataset;
(2) If we consider the relative performance improvement, i.e. the ratio of the results of the basic method to the enhanced method, we notice that the improvement is more significant on dataset IMDB than DBLP. Simultaneously, Figure \ref{fig:bias} presents that IMDB exhibits stronger topological bias, so debiasing on this dataset will better enhance the overall performance, which is consistent with the experimental results.
Table \ref{tab:debias} shows the performance comparison between HTAD and other GCL methods. It can be seen that in some cases, two general GCL methods can slightly improve the performance and partial heterogeneous GCL methods perform better. The proposed method outperforms all other baselines on all datasets and achieves state-of-the-art (SOTA) results, benefiting from the considieration of the topological bias and more effective edge sampling strategy.

\textbf{Debiasing performance measure.}
As we described in Section \ref{sec:bias}, topological bias is represented as the performance discrepancies of the model on each node. Thus we adopt variance as an indicator to measure the method's debiasing ability. A low variance indicates that the accuracy samples does not deviate too much from the mean, i.e. better efficacy. Specifically, we define two metrics, respectively total variance and bucket variance, which are respectively formulated as 
\begin{equation}
\begin{split}
    \mathrm{Var}&=\frac{1}{N}\sum_{i=1}^{N}(\text{acc}_i-\overline{\text{acc}})^2\\
    \mathrm{Var}_{b}&=\frac{1}{N_b}\sum_{i=1}^{N_b}\left(\overline{\text{acc}^{(i)}}-\overline{\text{acc}}\right)^2,
\end{split}
\end{equation}
where \(N_b\) is the number of buckets. \(\overline{\text{acc}}\) and \(\overline{\text{acc}^{(i)}}\) denote the average of accuracy on all nodes and nodes of bucket \(i\) respectively.
In experiments, we select projection function \(f=\mathrm{HLID}\) to construct buckets due to its superior property.
In Table \ref{tab:debias}, it's shown that HTAD generally outperforms the baselines on both two metrics in terms of debiasing capability, which indicates that the proposed method truly improves the model's performance on disadvantaged nodes, thereby enhancing the overall performance.

\begin{figure*}
    \centering
    \begin{subfigure}{0.18\linewidth}
        \centering
        \includegraphics[width=\linewidth]{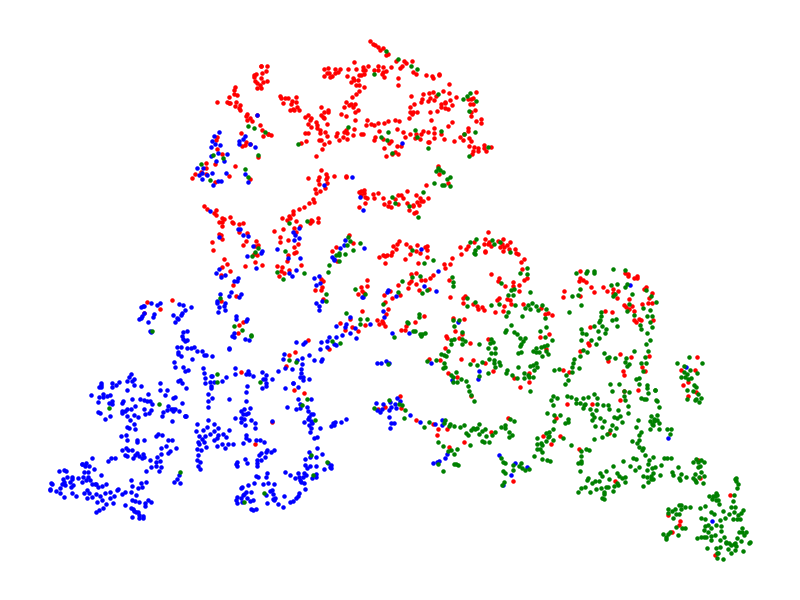}
        \caption{GCN}
    \end{subfigure}
    \hfill
    \begin{subfigure}{0.18\linewidth}
        \centering
        \includegraphics[width=\linewidth]{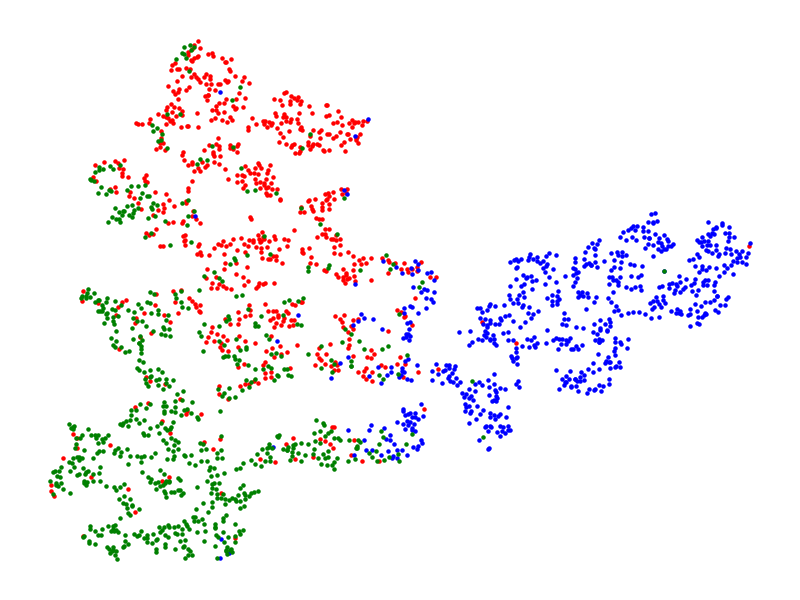}
        \caption{HAN}
    \end{subfigure}
    \hfill
    \begin{subfigure}{0.18\linewidth}
        \centering
        \includegraphics[width=\linewidth]{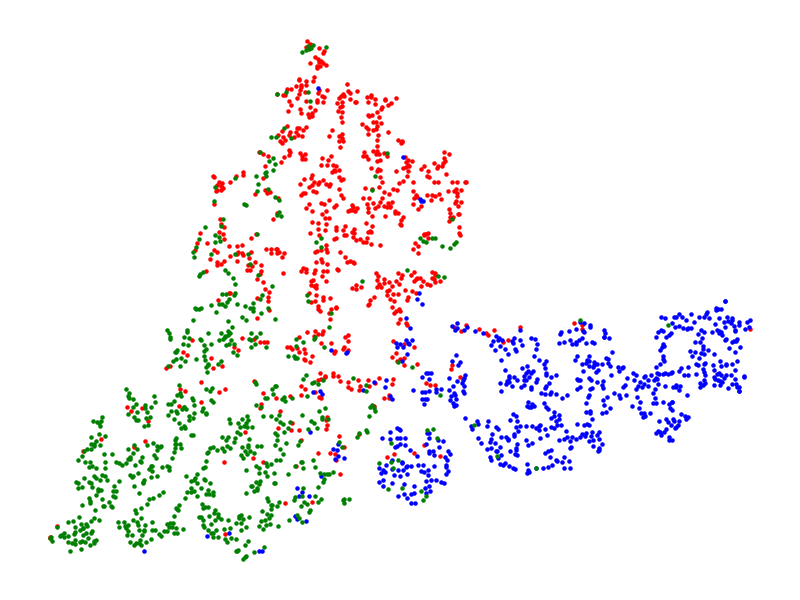}
        \caption{GCA}
    \end{subfigure}
    \hfill
    \begin{subfigure}{0.18\linewidth}
        \centering
        \includegraphics[width=\linewidth]{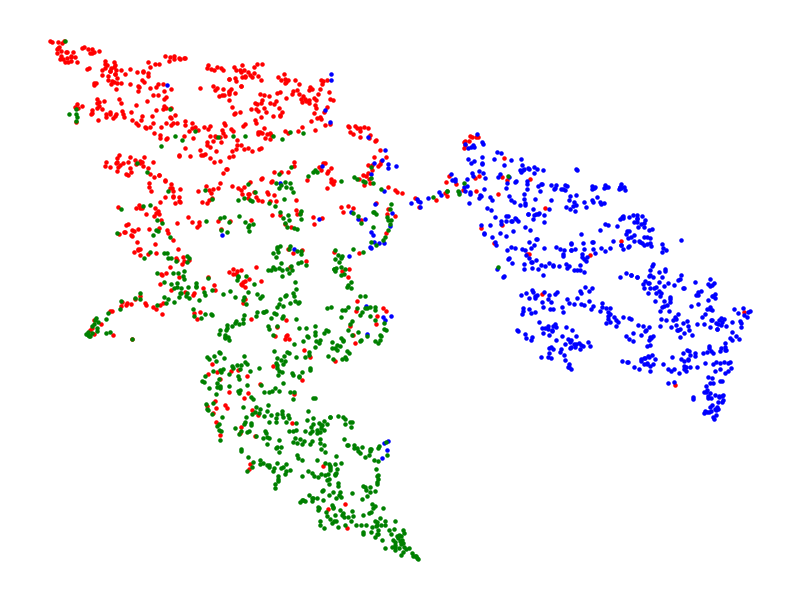}
        \caption{HGCL}
    \end{subfigure}
    \hfill
    \begin{subfigure}{0.18\linewidth}
        \centering
        \includegraphics[width=\linewidth]{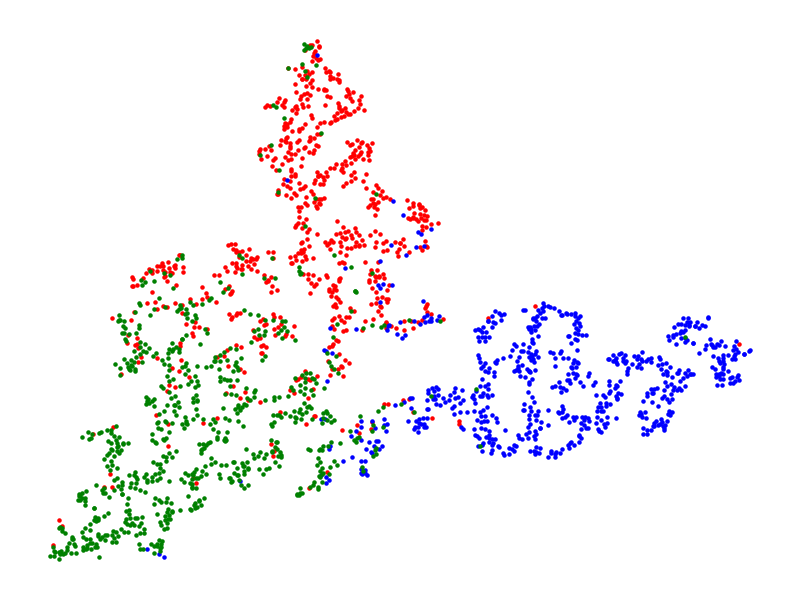}
        \caption{HTAD}
    \end{subfigure}
    \vspace{\baselineskip}
    
    \caption{Visualization of the learned node embedding on ACM. Three distinct categories are represented by three different colors.}
    \label{fig:tsne}
\end{figure*}

\textbf{Node clustering.}
In the node clustering task, we apply K-Means on the learned embeddings to obtain the cluster labels of nodes. As the results are not aligned with the true labels, we utilize widely-adopted normalized mutual information (NMI) and adjusted rand index (ARI) to quantify the performance.
Moreover, due to the particularity of the IMDB dataset (multi-label), we adopt fuzzy c-means (FCM) clustering \cite{bezdek1984fcm} to obtain a membership matrix where each element represents the probability of a node belonging to a cluster.
To evaluate the results of multi-label clustering, standard NMI and ARI are no longer applicable. For this purpose, inspired by \cite{campello2007fuzzy}, we calculate pairwise cosine similarity for membership vectors which results in a vector \(\overrightarrow{sim} \) of length \(\binom{n}{2}\). Then we calculate Manhattan (\(l_1\)) distance and Euclidean (\(l_2\)) distance between \(\overrightarrow{sim}_{\text{pred}}\) and \(\overrightarrow{sim}_{\text{true}}\). A smaller distance indicates that the predicted results are closer to the true value, i.e. the better the performance. The results are shown in Table \ref{tab:clus}.
Our method consistently outperforms existing methods, achieving significant improvement in different metrics on all datasets. This demonstrates that our approach effectively captures structural and feature information for better clustering.

\textbf{Qualitative analysis.}
To provide a more intuitive evaluation, we conduct qualitative analysis by visualizing the embeddings using t-SNE. We choose GCN, HAN, GCA and HGCL for comparison and the results are shown in Figure \ref{fig:tsne}, where different colors represent different labels.
It's demonstrated that that simply adopting the base model will result in the boundaries between regions with different labels very blurred. Benefiting from GCL, GCA and HGCL show slightly better performance, but there are still many different label areas overlapping. The proposed HTAD shows the best performance under t-SNE mapping, as the known categories form clear groupings and the clusters are compact and well-defined.

\subsection{Efficiency Analysis}
We evaluate the runtime efficiency of our method compared to baselines. To diminish the impact of randomness, we run 10 times and take the average. Table \ref{tab:eff} reports the training time per epoch (ms).
Overall, each method is fastest on ACM and slowest on DBLP, which is consistent with the scale of the dataset.
\begin{table}[]
\centering
\vspace{\baselineskip}
\caption{Training time per epoch. (ms)}
\vspace{\baselineskip}
\label{tab:eff}
\begin{tabular}{ccccccc}
\toprule
Dataset & GRACE  & GCA    & HeCo   & HGCL   & HTAD   \\ \midrule
ACM     & 7,653  & 8,714  & 12,842 & 8,787  & 7,686  \\
IMDB    & 13,416 & 16,406 & 27,092 & 16,849 & 13,470 \\
DBLP    & 32,337 & 37,142 & 86,134 & 40,837 & 32,887
\\ \bottomrule
\end{tabular}
\end{table}
The efficiency of GCA is negatively affected by a large number of redundant edges in the contrastive graph, while the efficiency bottlenecks of HeCo and HGCL lie in the computation of multiple meta-paths and the additional network overhead, respectively.
The proposed method achieves comparable efficiency to the simplest GRACE mothod by constructing sparse contrastive graph and simple framework.

\subsection{Parameter Sensitivity Analysis}
\begin{figure}[t]
    \centering
    \begin{subfigure}{0.48\linewidth}
        \centering
        \includegraphics[width=\linewidth]{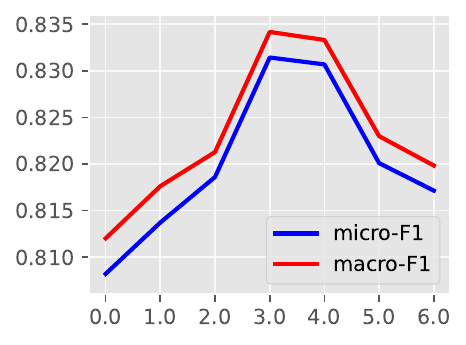}
        \caption{Effect on accuracy}
    \label{fig:eta1-acc}
    \end{subfigure}
    \hfill
    \begin{subfigure}{0.48\linewidth}
        \centering
        \includegraphics[width=\linewidth]{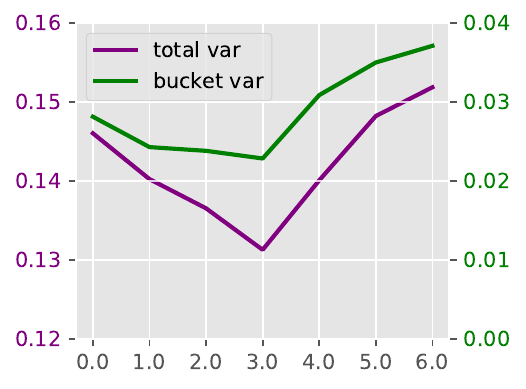}
        \caption{Effect on bias}
    \label{fig:eta1-var}
    \end{subfigure}
    \vspace{\baselineskip}
    \caption{Effect of regulator \(\eta_1\)}
    \vspace{\baselineskip}
    \label{fig:eta1}
\end{figure}
\begin{figure}[t]
    \centering
    \begin{subfigure}{0.48\linewidth}
        \centering
        \includegraphics[width=\linewidth]{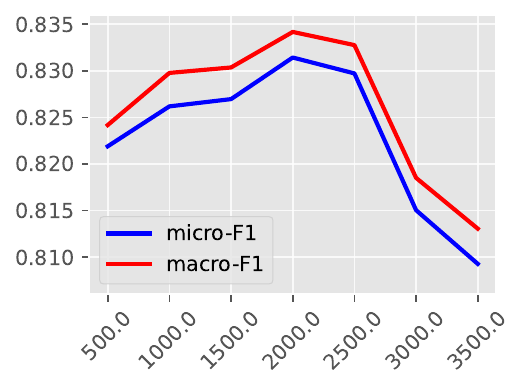}
        \caption{Effect on accuracy}
    \end{subfigure}
    \hfill
    \begin{subfigure}{0.48\linewidth}
        \centering
        \includegraphics[width=\linewidth]{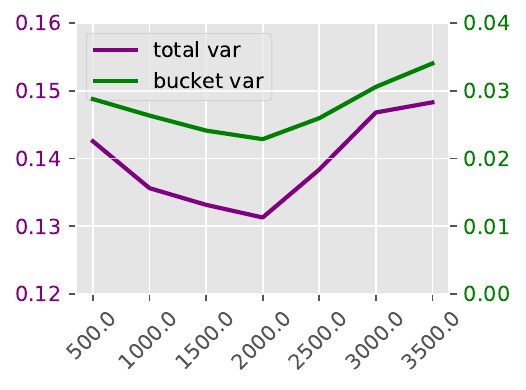}
        \caption{Effect on bias}
    \end{subfigure}
    \vspace{\baselineskip}
    \caption{Effect of regulator \(\eta_2\)}
    \vspace{1.5\baselineskip}
    \label{fig:eta2}
\end{figure}
\begin{figure}[t]
    \centering
    \begin{subfigure}{0.48\linewidth}
        \centering
        \includegraphics[width=\linewidth]{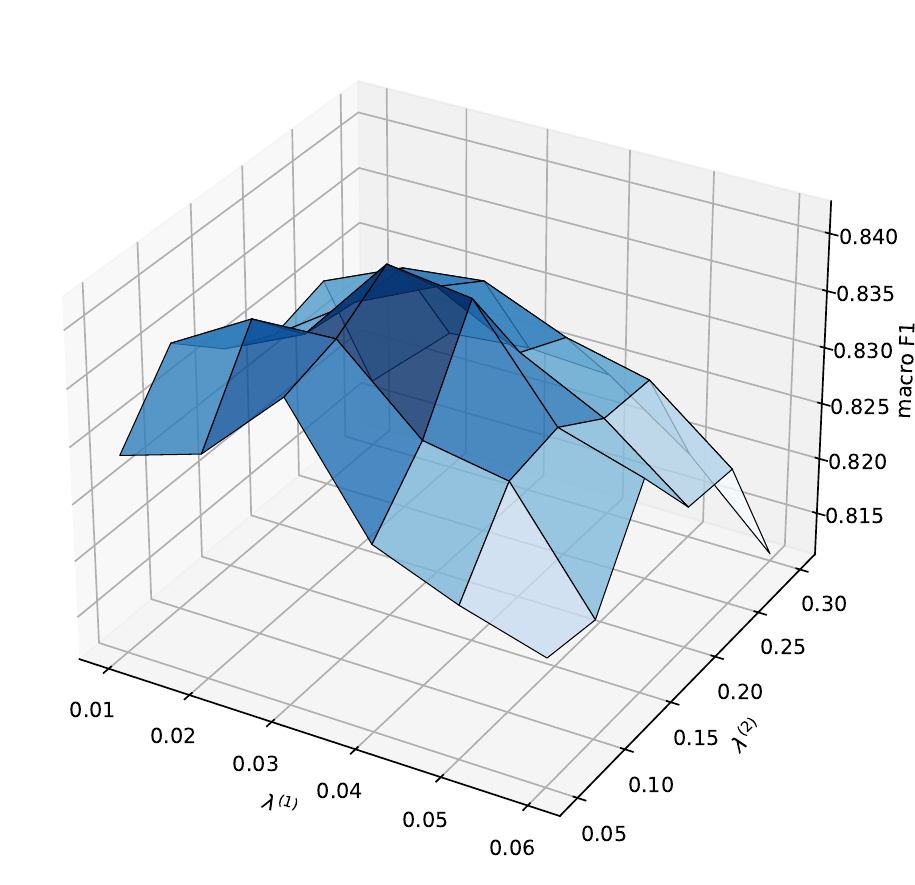}
        \caption{Effect on macro F1 score}
    \end{subfigure}
    \hfill
    \begin{subfigure}{0.48\linewidth}
        \centering
        \includegraphics[width=\linewidth]{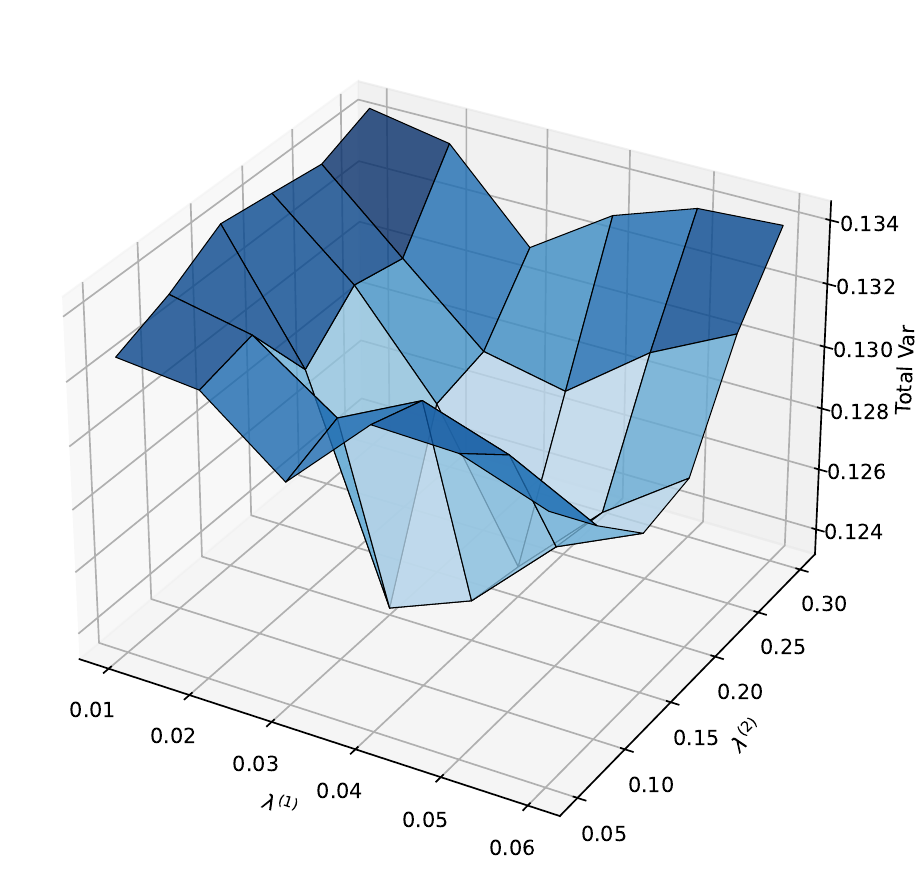}
        \caption{Effect on total variance}
    \end{subfigure}
    \vspace{\baselineskip}
    \caption{Effect of loss regulators}
    \vspace{\baselineskip}
    \label{fig:loss}
\end{figure}
In this subsection, we further study the parameter sensitivity of the proposed method.
Firstly, we study the impact of two regulators in Equation \ref{eq:mw}.
We set \(\eta_1=2000.0\) and vary \(\eta_2\), as shown in Figure \ref{fig:eta1}. The results demonstrate that the accuracy generally varies concavely with \(\eta_1\), peaking at \(\eta_1=3.0\) (Figure \ref{fig:eta1-acc}). When \(\eta_1\) is too small, the enhancement of intra-type connections is insufficient, and when it's too large, intra-type connections will dominate and impair other connections, which both resulting in reduced performance. The bucket variance curve follows a similar variation as the total variance and the model's debiasing performance is also the strongest when \(\eta_1=3.0\) (Figure \ref{fig:eta1-var}).
Similarly, we vary \(\eta_2\) while keeping \(\eta_1=3.0\) fixed and the results are presented in Figure \ref{fig:eta2} which shows similar trends.
The model can achieve optimal performance by adjusting the weights of two constructed terms, which indicates the strong correlation between them and strengthens the effectiveness of the proposed meta-weighting method.

We also test performance under different coefficients of contrastive losses and the results are shown in Figure \ref{fig:loss}. It's demonstrated that the accuracy roughly presenting a 2D concave surface, while the total variance is convex. The optimal solutions are both located in the same area (at around \(\lambda^{(1)}=0.3\) and \(\lambda^{(2)}=0.15\)). When the two coefficients are too large, the effect of label loss will be submerged, leading to suboptimal results.
Nevertheless, we notice that as the parameter combination changes, the method can still exceed the optimal results of most baselines.
It suggests that two losses both facilitate the model training, and demand balance of contributions.

\section{Conclusion}
In this paper, we study the topological bias in HGNNs and demonstrate its existence in datasets both with and without intra-type connections. The constructed projection, HLID, which is based on the meta-weighted graph, proves to be effective in revealing the inherent topological bias between nodes with different structural properties. We also propose a debiasing structure based on graph contrastive learning, which has been shown to enhance HGNNs' performance while mitigating bias in semi-supervised learning tasks.

We hope that this work will help address some of the current bottlenecks in HGNN research and contribute to the advancement of applications such as recommendation systems and beyond.






\bibliography{m7065}

\begin{thebibliography}{42}
\providecommand{\natexlab}[1]{#1}
\providecommand{\url}[1]{\texttt{#1}}
\expandafter\ifx\csname urlstyle\endcsname\relax
  \providecommand{\doi}[1]{doi: #1}\else
  \providecommand{\doi}{doi: \begingroup \urlstyle{rm}\Url}\fi

\bibitem[Albert and Barab{\'a}si(2002)]{albert2002statistical}
R.~Albert and A.-L. Barab{\'a}si.
\newblock Statistical mechanics of complex networks.
\newblock \emph{Reviews of Modern Physics}, 74\penalty0 (1):\penalty0 47, 2002.

\bibitem[Bachman et~al.(2019)Bachman, Hjelm, and Buchwalter]{bachman2019learning}
P.~Bachman, R.~D. Hjelm, and W.~Buchwalter.
\newblock Learning representations by maximizing mutual information across views.
\newblock \emph{Advances in Neural Information Processing Systems}, 32, 2019.

\bibitem[Barab{\'a}si and Albert(1999)]{barabasi1999emergence}
A.-L. Barab{\'a}si and R.~Albert.
\newblock Emergence of scaling in random networks.
\newblock \emph{Science}, 286\penalty0 (5439):\penalty0 509--512, 1999.

\bibitem[Bezdek et~al.(1984)Bezdek, Ehrlich, and Full]{bezdek1984fcm}
J.~C. Bezdek, R.~Ehrlich, and W.~Full.
\newblock Fcm: The fuzzy c-means clustering algorithm.
\newblock \emph{Computers \& Geosciences}, 10\penalty0 (2-3):\penalty0 191--203, 1984.

\bibitem[Cai et~al.(2021)Cai, Luo, Xu, He, Liu, and Wang]{cai2021graphnorm}
T.~Cai, S.~Luo, K.~Xu, D.~He, T.-y. Liu, and L.~Wang.
\newblock Graphnorm: A principled approach to accelerating graph neural network training.
\newblock In \emph{International Conference on Machine Learning}, pages 1204--1215. PMLR, 2021.

\bibitem[Campello(2007)]{campello2007fuzzy}
R.~J. Campello.
\newblock A fuzzy extension of the rand index and other related indexes for clustering and classification assessment.
\newblock \emph{Pattern Recognition Letters}, 28\penalty0 (7):\penalty0 833--841, 2007.

\bibitem[Che et~al.(2021)Che, Tao, Yang, Liu, and Zhang]{che2021multi}
F.~Che, J.~Tao, G.~Yang, T.~Liu, and D.~Zhang.
\newblock Multi-aspect self-supervised learning for heterogeneous information network.
\newblock \emph{Knowledge-Based Systems}, 233:\penalty0 107474, 2021.

\bibitem[Chen et~al.(2023)Chen, Huang, Xia, Wei, Xu, and Luo]{chen2023heterogeneous}
M.~Chen, C.~Huang, L.~Xia, W.~Wei, Y.~Xu, and R.~Luo.
\newblock Heterogeneous graph contrastive learning for recommendation.
\newblock In \emph{Proceedings of the Sixteenth ACM International Conference on Web Search and Data Mining}, pages 544--552, 2023.

\bibitem[Chen et~al.(2020)Chen, Kornblith, Norouzi, and Hinton]{chen2020simple}
T.~Chen, S.~Kornblith, M.~Norouzi, and G.~Hinton.
\newblock A simple framework for contrastive learning of visual representations.
\newblock In \emph{International Conference on Machine Learning}, pages 1597--1607. PmLR, 2020.

\bibitem[Dong et~al.(2017)Dong, Chawla, and Swami]{dong2017metapath2vec}
Y.~Dong, N.~V. Chawla, and A.~Swami.
\newblock metapath2vec: Scalable representation learning for heterogeneous networks.
\newblock In \emph{Proceedings of the 23rd ACM SIGKDD International Conference on Knowledge Discovery and Data Mining}, pages 135--144, 2017.

\bibitem[Fu et~al.(2020)Fu, Zhang, Meng, and King]{fu2020magnn}
X.~Fu, J.~Zhang, Z.~Meng, and I.~King.
\newblock Magnn: Metapath aggregated graph neural network for heterogeneous graph embedding.
\newblock In \emph{Proceedings of The Web Conference 2020}, pages 2331--2341, 2020.

\bibitem[Gasteiger et~al.(2018)Gasteiger, Bojchevski, and G{\"u}nnemann]{gasteiger2018predict}
J.~Gasteiger, A.~Bojchevski, and S.~G{\"u}nnemann.
\newblock Predict then propagate: Graph neural networks meet personalized pagerank.
\newblock In \emph{International Conference on Learning Representations}, 2018.

\bibitem[Hamilton et~al.(2017)Hamilton, Ying, and Leskovec]{hamilton2017inductive}
W.~Hamilton, Z.~Ying, and J.~Leskovec.
\newblock Inductive representation learning on large graphs.
\newblock \emph{Advances in Neural Information Processing Systems}, 30, 2017.

\bibitem[Han et~al.(2024)Han, Liu, Shi, Torkamani, Aggarwal, and Tang]{han2024towards}
H.~Han, X.~Liu, F.~Shi, M.~Torkamani, C.~Aggarwal, and J.~Tang.
\newblock Towards label position bias in graph neural networks.
\newblock \emph{Advances in Neural Information Processing Systems}, 36, 2024.

\bibitem[Hassani and Khasahmadi(2020)]{hassani2020contrastive}
K.~Hassani and A.~H. Khasahmadi.
\newblock Contrastive multi-view representation learning on graphs.
\newblock In \emph{International Conference on Machine Learning}, pages 4116--4126. PMLR, 2020.

\bibitem[Hjelm et~al.(2019)Hjelm, Fedorov, Lavoie-Marchildon, Grewal, Bachman, Trischler, and Bengio]{hjelm2018learning}
R.~D. Hjelm, A.~Fedorov, S.~Lavoie-Marchildon, K.~Grewal, P.~Bachman, A.~Trischler, and Y.~Bengio.
\newblock Learning deep representations by mutual information estimation and maximization.
\newblock In \emph{International Conference on Learning Representations}, 2019.
\newblock URL \url{https://openreview.net/forum?id=Bklr3j0cKX}.

\bibitem[Hu et~al.(2020)Hu, Dong, Wang, and Sun]{hu2020heterogeneous}
Z.~Hu, Y.~Dong, K.~Wang, and Y.~Sun.
\newblock Heterogeneous graph transformer.
\newblock In \emph{Proceedings of The Web Conference 2020}, pages 2704--2710, 2020.

\bibitem[Jiang et~al.(2021{\natexlab{a}})Jiang, Jia, Fang, Shi, Lin, and Wang]{jiang2021pre}
X.~Jiang, T.~Jia, Y.~Fang, C.~Shi, Z.~Lin, and H.~Wang.
\newblock Pre-training on large-scale heterogeneous graph.
\newblock In \emph{Proceedings of the 27th ACM SIGKDD Conference on Knowledge Discovery \& Data Mining}, pages 756--766, 2021{\natexlab{a}}.

\bibitem[Jiang et~al.(2021{\natexlab{b}})Jiang, Lu, Fang, and Shi]{jiang2021contrastive}
X.~Jiang, Y.~Lu, Y.~Fang, and C.~Shi.
\newblock Contrastive pre-training of gnns on heterogeneous graphs.
\newblock In \emph{Proceedings of the 30th ACM International Conference on Information \& Knowledge Management}, pages 803--812, 2021{\natexlab{b}}.

\bibitem[Kingma and Ba(2014)]{kingma2014adam}
D.~P. Kingma and J.~Ba.
\newblock Adam: A method for stochastic optimization.
\newblock \emph{arXiv preprint arXiv:1412.6980}, 2014.

\bibitem[Kipf and Welling(2017)]{kipf2017semi}
T.~N. Kipf and M.~Welling.
\newblock Semi-supervised classification with graph convolutional networks.
\newblock In \emph{International Conference on Learning Representations}, 2017.

\bibitem[Lv et~al.(2021)Lv, Ding, Liu, Chen, Feng, He, Zhou, Jiang, Dong, and Tang]{lv2021we}
Q.~Lv, M.~Ding, Q.~Liu, Y.~Chen, W.~Feng, S.~He, C.~Zhou, J.~Jiang, Y.~Dong, and J.~Tang.
\newblock Are we really making much progress? revisiting, benchmarking and refining heterogeneous graph neural networks.
\newblock In \emph{Proceedings of the 27th ACM SIGKDD Conference on Knowledge Discovery \& Data Mining}, pages 1150--1160, 2021.

\bibitem[Park et~al.(2020)Park, Kim, Han, and Yu]{park2020unsupervised}
C.~Park, D.~Kim, J.~Han, and H.~Yu.
\newblock Unsupervised attributed multiplex network embedding.
\newblock In \emph{Proceedings of the AAAI Conference on Artificial Intelligence}, volume~34, pages 5371--5378, 2020.

\bibitem[Sohn(2016)]{sohn2016improved}
K.~Sohn.
\newblock Improved deep metric learning with multi-class n-pair loss objective.
\newblock \emph{Advances in Neural Information Processing Systems}, 29, 2016.

\bibitem[Spearman(1987)]{spearman1987proof}
C.~Spearman.
\newblock The proof and measurement of association between two things.
\newblock \emph{The American Journal of Psychology}, 100\penalty0 (3/4):\penalty0 441--471, 1987.

\bibitem[Tang et~al.(2020)Tang, Yao, Sun, Wang, Tang, Aggarwal, Mitra, and Wang]{tang2020investigating}
X.~Tang, H.~Yao, Y.~Sun, Y.~Wang, J.~Tang, C.~Aggarwal, P.~Mitra, and S.~Wang.
\newblock Investigating and mitigating degree-related biases in graph convoltuional networks.
\newblock In \emph{Proceedings of the 29th ACM International Conference on Information \& Knowledge Management}, pages 1435--1444, 2020.

\bibitem[Van Der~Hofstad(2014)]{van2014random}
R.~Van Der~Hofstad.
\newblock Random graphs and complex networks.
\newblock \emph{vol. I}, 2014.

\bibitem[Veli{\v{c}}kovi{\'c} et~al.(2018)Veli{\v{c}}kovi{\'c}, Cucurull, Casanova, Romero, Li{\`o}, and Bengio]{velivckovic2018graph}
P.~Veli{\v{c}}kovi{\'c}, G.~Cucurull, A.~Casanova, A.~Romero, P.~Li{\`o}, and Y.~Bengio.
\newblock Graph attention networks.
\newblock In \emph{International Conference on Learning Representations}, 2018.

\bibitem[Veli{\v{c}}kovi{\'c} et~al.(2019)Veli{\v{c}}kovi{\'c}, Fedus, Hamilton, Li{\`o}, Bengio, and Hjelm]{velivckovic2018deep}
P.~Veli{\v{c}}kovi{\'c}, W.~Fedus, W.~L. Hamilton, P.~Li{\`o}, Y.~Bengio, and R.~D. Hjelm.
\newblock Deep graph infomax.
\newblock In \emph{International Conference on Learning Representations}, 2019.

\bibitem[Wan et~al.(2021)Wan, Pan, Yang, and Gong]{wan2021contrastive}
S.~Wan, S.~Pan, J.~Yang, and C.~Gong.
\newblock Contrastive and generative graph convolutional networks for graph-based semi-supervised learning.
\newblock In \emph{Proceedings of the AAAI Conference on Artificial Intelligence}, volume~35, pages 10049--10057, 2021.

\bibitem[Wang et~al.(2022)Wang, Wang, Shi, and Song]{wang2022uncovering}
R.~Wang, X.~Wang, C.~Shi, and L.~Song.
\newblock Uncovering the structural fairness in graph contrastive learning.
\newblock \emph{Advances in Neural Information Processing Systems}, 35:\penalty0 32465--32473, 2022.

\bibitem[Wang et~al.(2019)Wang, Ji, Shi, Wang, Ye, Cui, and Yu]{wang2019heterogeneous}
X.~Wang, H.~Ji, C.~Shi, B.~Wang, Y.~Ye, P.~Cui, and P.~S. Yu.
\newblock Heterogeneous graph attention network.
\newblock In \emph{The World Wide Web Conference}, pages 2022--2032, 2019.

\bibitem[Wang et~al.(2021)Wang, Liu, Han, and Shi]{wang2021self}
X.~Wang, N.~Liu, H.~Han, and C.~Shi.
\newblock Self-supervised heterogeneous graph neural network with co-contrastive learning.
\newblock In \emph{Proceedings of the 27th ACM SIGKDD Conference on Knowledge Discovery \& Data Mining}, pages 1726--1736, 2021.

\bibitem[Xu et~al.(2018)Xu, Hu, Leskovec, and Jegelka]{xu2018powerful}
K.~Xu, W.~Hu, J.~Leskovec, and S.~Jegelka.
\newblock How powerful are graph neural networks?
\newblock In \emph{International Conference on Learning Representations}, 2018.

\bibitem[Yang et~al.(2023)Yang, Yan, Pan, Ye, and Fan]{yang2023simple}
X.~Yang, M.~Yan, S.~Pan, X.~Ye, and D.~Fan.
\newblock Simple and efficient heterogeneous graph neural network.
\newblock In \emph{Proceedings of the AAAI Conference on Artificial Intelligence}, volume~37, pages 10816--10824, 2023.

\bibitem[Yao et~al.(2023)Yao, Wang, Li, Zhu, Jiang, Li, Tian, Yang, Liu, and Liu]{yao2023semi}
K.~Yao, X.~Wang, W.~Li, H.~Zhu, Y.~Jiang, Y.~Li, T.~Tian, Z.~Yang, Q.~Liu, and Q.~Liu.
\newblock Semi-supervised heterogeneous graph contrastive learning for drug--target interaction prediction.
\newblock \emph{Computers in Biology and Medicine}, 163:\penalty0 107199, 2023.

\bibitem[You et~al.(2020)You, Chen, Sui, Chen, Wang, and Shen]{you2020graph}
Y.~You, T.~Chen, Y.~Sui, T.~Chen, Z.~Wang, and Y.~Shen.
\newblock Graph contrastive learning with augmentations.
\newblock \emph{Advances in Neural Information Processing Systems}, 33:\penalty0 5812--5823, 2020.

\bibitem[Yu et~al.(2024)Yu, Ge, Li, and Zhou]{yu2024heterogeneous}
J.~Yu, Q.~Ge, X.~Li, and A.~Zhou.
\newblock Heterogeneous graph contrastive learning with meta-path contexts and adaptively weighted negative samples.
\newblock \emph{IEEE Transactions on Knowledge and Data Engineering}, 2024.

\bibitem[Zhu et~al.(2020)Zhu, Xu, Yu, Liu, Wu, and Wang]{zhu2020deep}
Y.~Zhu, Y.~Xu, F.~Yu, Q.~Liu, S.~Wu, and L.~Wang.
\newblock Deep graph contrastive representation learning.
\newblock In \emph{ICML 2020 Workshop}, 2020.
\newblock URL \url{https://grlplus.github.io/papers/12.pdf}.

\bibitem[Zhu et~al.(2021{\natexlab{a}})Zhu, Xu, Liu, and Wu]{zhu2021an}
Y.~Zhu, Y.~Xu, Q.~Liu, and S.~Wu.
\newblock An empirical study of graph contrastive learning.
\newblock In \emph{Thirty-fifth Conference on Neural Information Processing Systems Datasets and Benchmarks Track (Round 2)}, 2021{\natexlab{a}}.
\newblock URL \url{https://openreview.net/forum?id=UuUbIYnHKO}.

\bibitem[Zhu et~al.(2021{\natexlab{b}})Zhu, Xu, Yu, Liu, Wu, and Wang]{zhu2021graph}
Y.~Zhu, Y.~Xu, F.~Yu, Q.~Liu, S.~Wu, and L.~Wang.
\newblock Graph contrastive learning with adaptive augmentation.
\newblock In \emph{Proceedings of the Web Conference 2021}, pages 2069--2080, 2021{\natexlab{b}}.

\bibitem[Zhu et~al.(2022)Zhu, Xu, Cui, Yang, Liu, and Wu]{zhu2022structure}
Y.~Zhu, Y.~Xu, H.~Cui, C.~Yang, Q.~Liu, and S.~Wu.
\newblock Structure-enhanced heterogeneous graph contrastive learning.
\newblock In \emph{Proceedings of the 2022 SIAM International Conference on Data Mining (SDM)}, pages 82--90. SIAM, 2022.

\end{thebibliography}

\end{document}